%% file: main.tex
\documentclass[11pt, a4paper, logo, twocolumn]{googledeepmind}

\usepackage{subcaption}
\usepackage[capitalize,noabbrev]{cleveref}
\usepackage{animate}

\usepackage[authoryear, sort&compress, round]{natbib}
\input{preamble}

\title{Direct Motion Models for Assessing Generated Videos}

\correspondingauthor{krallen@google.com,svansteenkiste@google.com}

\author[*,1]{Kelsey Allen}
\author[1]{Carl Doersch}
\author[1]{Guangyao Zhou}
\author[2]{Mohammed Suhail}
\author[1]{Danny Driess}
\author[1]{Ignacio Rocco}
\author[1]{Yulia Rubanova}
\author[1]{Thomas Kipf}
\author[1]{Mehdi S. M. Sajjadi}
\author[1]{Kevin Murphy}
\author[1]{Joao Carreira}
\author[*,2]{Sjoerd van Steenkiste}

\affil[*]{Equal contributions}
\affil[1]{Google DeepMind}
\affil[2]{Google Research}

\input{sections/0_abstract}

\begin{document}

\maketitle

\input{sections/1_introduction}

\input{sections/2_related_work}

\input{sections/3_method}
\input{sections/5_results}

\input{sections/6_conclusion}

\bibliography{main}

\newpage
\appendix
\onecolumn
\input{sections/X_suppl}

\end{document}

%% file: preamble.tex
\usepackage{xcolor}         
\usepackage{xspace}

\usepackage{multirow}
\usepackage{wrapfig}
\usepackage{enumitem}

\makeatletter
\DeclareRobustCommand\onedot{\futurelet\@let@token\@onedot}
\def\@onedot{\ifx\@let@token.\else.\null\fi\xspace}

\def\eg{\emph{e.g}\onedot}

\newcommand{\model}{TRAJAN\xspace}



%% file: sections/0_abstract.tex
\begin{abstract}
A current limitation of video generative video models is that they generate plausible looking frames, but poor motion --- an issue that is not well captured by FVD and other popular methods for evaluating generated videos.
Here we go beyond FVD by developing a metric which better measures plausible object interactions and motion.
Our novel approach is based on auto-encoding point tracks and yields motion features that can be used to not only compare distributions of videos (as few as one generated and one ground truth, or as many as two datasets), but also for evaluating motion of single videos.
We show that using point tracks instead of pixel reconstruction or action recognition features results in a metric which is markedly more sensitive to temporal distortions in synthetic data, and can predict human evaluations of temporal consistency and realism in generated videos obtained from open-source models better than a wide range of alternatives. We also show that by using a point track representation, we can spatiotemporally localize generative video inconsistencies, providing extra interpretability of generated video errors relative to prior work. An overview of the results and link to the code can be found on the project page: \url{trajan-paper.github.io}.\looseness=-1
\end{abstract}

%% file: sections/1_introduction.tex
\section{Introduction}
\label{sec:intro}

As generative video models become increasingly capable, the community needs more powerful methods for automatically evaluating the quality of generated videos.
State-of-the-art models are getting better at generating what appear to be plausible looking frames, yet they still struggle to put together coherent motion~\citep{videoworldsimulators2024}. 
 The current gold standard for assessing video quality is collecting human judgments, but these are expensive to obtain and not scalable as a metric for regularly measuring improvements in modeling capabilities, e.g.\ throughout training.

Existing metrics, such as those based on Fr\'{e}chet Video Distance (FVD)~\citep{unterthiner2018towards,ge2024content,luo2024beyond}, can capture certain elements of plausibility, but are more sensitive to frame-level content effects~\citep{ge2024content}, and depend on access to the underlying training distribution which is not always available.
Further, these approaches cover only one way of evaluating generated videos, i.e.\ by comparing entire distributions, and it remains unclear how to evaluate pairs of videos, or individual videos, with this approach. 

Here we propose a new method that we show addresses many of these issues, by directly modelling 2D video motion.
Given a single generated video, we estimate low-level, temporally-extended motion features as point tracks using the publicly available BootsTAPIR model~\citep{doersch2024bootstap}.
Next, we (auto)encode these features to obtain dense high-level motion features using a novel \emph{TRAJectory AutoeNcoder (\model)} architecture. 
We can then use the \model latent space to compare distributions of videos (as few as one generated and one real, or as many as two datasets), or the reconstruction error from \model to estimate per-video motion inconsistencies.

\begin{figure}
    \centering
    \begin{subfigure}{\columnwidth}
    \includegraphics[width=\columnwidth]{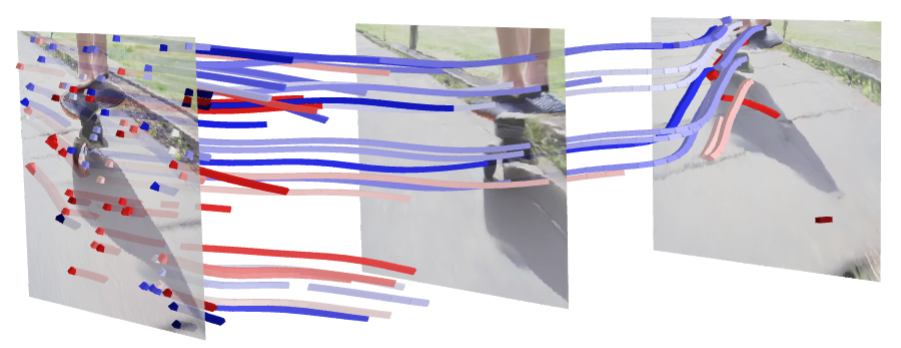}
    \caption{Unrealistic object appearance, but plausible motion.}
    \end{subfigure}
    \begin{subfigure}{\columnwidth}
    \includegraphics[width=\columnwidth]{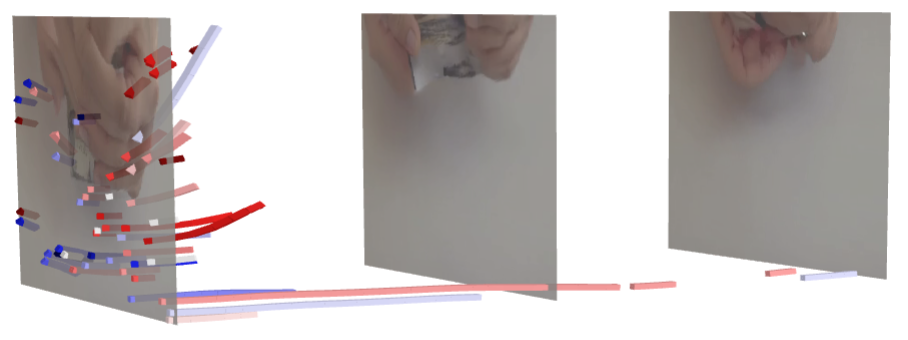}
    \caption{Realistic looking hands, but which change appearance from frame-to-frame, with fingers morphing in and out of existence.}
    \end{subfigure}
    \caption{Predicted point trajectories from \model, colored by their reconstruction error with respect to BootsTAPIR (\textcolor{blue}{well reconstructed track}, \textcolor{red}{poorly reconstructed track})~\citep{doersch2024bootstap}. See \url{trajan-paper.github.io} for videos. The video in (a) has plausible motion but implausible frame-level appearance, while the video in (b) has plausible frame-level appearance but implausible motion (the fingers morph and disappear between frames). \model correctly predicts this discrepancy by focusing on \emph{motion} irrespective of \emph{appearance}. \vspace{-2em}}
    \label{fig:figintro}
\end{figure}

Point tracking, by design, separates the semantics from the motion content by focusing on features necessary for predicting motion without reconstructing the whole scene (\autoref{fig:figintro}).
This means that it will likely focus on plausible motion irrespective of semantic information.
We show that this is indeed the case across the three major ways in which the community generally uses metrics to evaluate videos:

\begin{enumerate}
    \item \textbf{At the distribution level}, we show that among several other choices --- including VideoMAE v2~\citep{wang2023videomae}, I3D~\citep{carreira2017quo}, and motion histograms~\citep{liu2024frechet} --- the features learned by TRAJAN are markedly more sensitive for detecting synthethic temporal distortions  (\S\ref{sec:results-distributional}).
    \item \textbf{For pairs of videos}, we compute a distance between their TRAJAN embeddings (\S\ref{sec:video-to-video}), which better captures similarities in motion even when appearance-based pixel-level metrics suggest that they are different.
    \item \textbf{For individual videos}, we measure how well TRAJAN is able to reconstruct the original input tracks (\S\ref{sec:results-per-video}). We find that this reconstruction score is more or equally predictive of human ratings for realism, appearance, and motion consistency of generated videos from 10 different open-source models than a wide range of alternatives (both appearance- and motion-based).
\end{enumerate}

\vspace{0.5em}
As an additional contribution, we conduct a detailed human study on generated videos. We collect a new dataset of ratings that focuses on four aspects of motion: quality of interactions, the realism of the objects and behavior, consistency of appearance and motion, and the speed of objects and the camera.
While acknowledging these are somewhat subjective, we find that \model typically outperforms prior methods at predicting human ratings, although with room for improvement, suggesting an area for further research.

%% file: sections/2_related_work.tex
\section{Related Work}
Metrics for assessing various aspects of video generation quality can broadly be categorized into a) comparing a distribution of generated videos with a reference distribution, b) comparing a generated video to a real ``ground truth'' reference, and c) operating on individual videos without a reference.
Though these areas are generally considered separately, here we propose \model as a unifying model for measuring motion consistency in all of these three settings.

\textbf{Metrics for video distributions.}
FVD \citep{unterthiner2018towards} is a standard metric to compare a distribution of real world videos with generated ones \citep{bugliarello2024storybench}.
However, \citet{ge2024content} show that FVD is biased towards the content of individual frames, possibly because of how the underlying I3D \citep{carreira2017quo} feature extractor was trained.
\citet{ge2024content} propose to use VideoMAE features to address this, while concurrent work by \citet{luo2024beyond} considers a number of alternative appearance-based feature representations. 

\citet{liu2024frechet} also propose a new feature representation, but they compute histogram-based motion features derived from estimated point tracks.
Our approach is closely related in that we also use point tracks, however, a key difference is that we use learned TRAJAN features, which we find to perform markedly better. 
Additionally, we demonstrate how TRAJAN can be used to assess the quality of individual videos.

\textbf{Metrics for paired videos.}
Evaluating the quality of generated video against ground truth frames often involves pixel-wise and perceptual metrics, such as Peak Signal-to-Noise Ratio (PSNR), Structural Similarity Index (SSIM), and Learned Perceptual Image Patch Similarity (LPIPS). PSNR measures pixel-level discrepancies using Mean Squared Error, but its pixel dependency makes it overly sensitive to small, often imperceptible variations. SSIM~\citep{wang2004image} improves on this by comparing local patterns in brightness, contrast, and structure to better align with human perception. LPIPS~\citep{zhang2018perceptual} uses a pre-trained deep networks (\eg, VGG~\citep{simonyan2014very}, AlexNet~\citep{krizhevsky2012imagenet}) to compute similarity based on feature maps, capturing higher-level perceptual cues. While these metrics provide valuable insights into frame-level similarity, they evaluate each frame independently and thus fail to capture temporal coherence—an essential component of video realism. 
\begin{figure*}
    \captionsetup[subfigure]{justification=centering}
    \centering
    \begin{subfigure}[t]{0.5\textwidth}
        \centering
        \includegraphics[width=\textwidth]{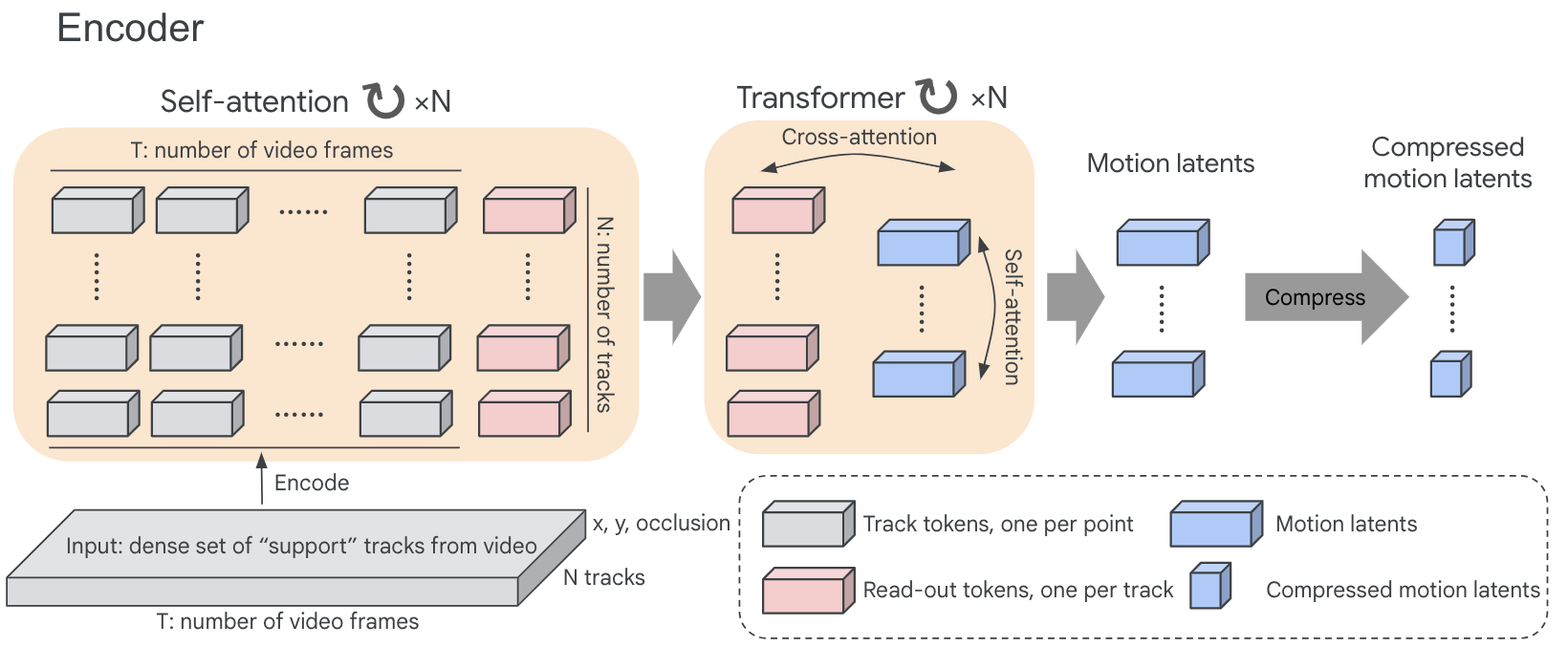}
        \caption{Point trajectory encoder}
    \end{subfigure}%
    ~ 
    \begin{subfigure}[t]{0.5\textwidth}
        \centering
        \includegraphics[width=\textwidth]{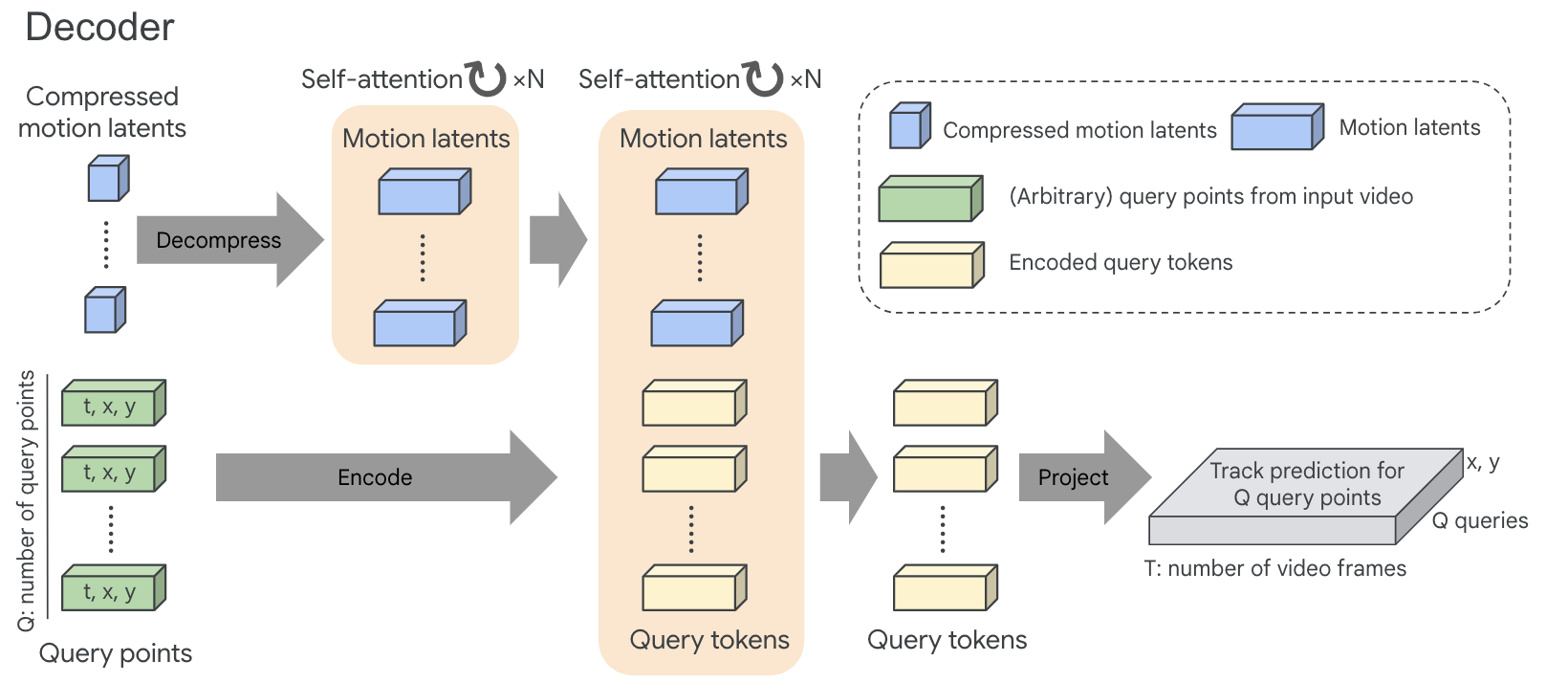}
        \caption{Query point decoder}
    \end{subfigure}
    \caption{\textbf{\model} -- The trajectory encoder encodes a (variable-sized) set of point trajectories $(x_{t,j},y_{t,j},o_{t,j},t)$ into a compressed motion latent $\phi_S$ of fixed size using a Perceiver~\citep{jaegle2021perceiver}-style transformer architecture. The occlusion flag $o$ is used in the attention mask, making the representation invariant to occluded points. The decoder takes this latent $\phi_S$ and predicts for a query point $(x_q, y_q, t_q)$ the point track that goes through this point at all other times, as well as their occlusion flag. By training the autoencoder on different input and query points, the model learns to represent a dense motion field.}
    \label{fig:trackautoencoder}
\end{figure*}

\textbf{Metrics for individual videos.}
Several recent metrics have been proposed for evaluating text-to-video models on a per-video basis, since in this case a reference video is typically unavailable.
However, a major focus has been on the semantic alignment between the prompt and its generated video.
CLIPScore \citep{hessel2021clipscore} utilizes CLIP \citep{radford2021learning} to compare individual video frames with the text in the embedding space \citep{gu2024videoswap, liu2024dynvideo, wu2023tune}.
The main issue with CLIPScore is that it mainly captures semantics of individual frames, instead of dynamic properties of videos such as consistency over time, physical plausibility, or motion quality \citep{bansal2024videophy}.
Even for image generation tasks, it has been shown that such metrics are not always aligned with human judgement \citep{otani2023toward}.
Beyond CLIP, other approaches propose to use vision-language models (VLMs) to judge the semantic adherence between the prompt and the video \citep{wu2024towards, bansal2024videophy}, either zero-shot or by finetuning a VLM on human annotations of video quality \citep{huang2024vbench}.
DOVER \citep{wu2022disentangling} is a model trained to predict the average human subjective perception quality of a video.
In addition to semantic adherence, VideoPhy by \citet{bansal2024videophy} evaluate videos by whether they follow physical commonsense by finetuning a VLM
\citep{bansal2024videocon} on this task.

Compared to these metrics, our approach is \emph{not} focused on semantic adherence between a prompt and a video, but rather on the motion and appearance consistency of a video \emph{independent of the prompt}. Our approach based on point tracks inherently takes the temporal evolution of the frames into account, which we show is crucial for matching human judgements of video quality on a per-video basis.

\textbf{Benchmarks \& generative video models.}
The field of generative video models is rapidly evolving.
Most current methods for text-to-video models are either based on diffusion \citep{he2022latent, ho2022video, ho2022imagen, singer2022make, luo2023videofusion, wang2023modelscope, zhou2022magicvideo, khachatryan2023text2video, blattmann2023align} or transformers \citep{villegas2022phenaki, wu2021godiva, hong2022cogvideo, wu2022nuwa}.
Hence, there have been several efforts to evaluate such generative models \citep{liu2024fetv, huang2024vbench}.
VideoPhy \citep{bansal2024videophy} is a benchmark that focuses on prompts where a model has to obey the laws of physics, such as marbles rolling down a slanted surface.
We evaluate \model both on VideoPhy \citep{bansal2024videophy} and EvalCrafter \citep{liu2024evalcrafter}.

%% file: sections/3_method.tex
\section{Methods}
To obtain metrics for evaluating motion in videos, we focus on two aspects: how to extract latent representations from videos, and how to compute ordinal-valued metrics for videos that can directly be understood as ranking individual videos as being better or worse.
We propose \model, which gives us latent representations and can be used directly as a metric on a per-video basis.

For distribution-level comparisons, we compute the Fr\'echet distance between the latents in two datasets $P_R$ and $P_G$ as in prior work~\citep{unterthiner2018towards}. It is defined as $d(P_R, P_G) = min_{X,Y}E|X-Y|^2$ which simplifies to $d(P_R, P_G) = |\mu_R - \mu_G|^2 + Tr(\sum_R +\sum_G - 2(\Sigma_R\Sigma_G)^{\frac{1}{2}})$ when $P_R$ and $P_G$ are multivariate Gaussians. 
For video pair comparisons, we simply take the $L_2$ distance between the latents.
For single video evaluations, we use the ordinal number (such as reconstruction error) to determine quality.

\subsection{\model}
\label{sec:method}

Point tracking models like PIPs++~\citep{zheng2023pointodyssey} or BootsTAPIR~\citep{doersch2024bootstap} can track points in arbitrary videos surprisingly well.
Point tracks are inherently linked to \emph{motion} rather than \emph{appearance}, making them a strong candidate for assessing motion quality in videos.
However, a set of point tracks is inconvenient to work with because the tracks are orderless, making it difficult to compare two sets without a ground-truth reference.
Further, we must also contend with missing data due to occlusions.

To address this problem, we introduce the TRAJectory AutoeNcoder (\model, \autoref{fig:trackautoencoder}). At a high level, \model is trained to reconstruct point tracks starting from random queries across space and time, and can provide both a latent representation and a reconstruction score. \model operates on a set of point trajectories $S=\{s_{t,j}\}$ coming from, e.g., BootsTAPIR, where $s_{t,j}= (x_{t,j},y_{t,j},o_{t,j})$ corresponds to $x$ and $y$ positions and occlusion flag $o$ at time $t$ for the $j$th trajectory, and is trained to reconstruct a separate set of query trajectories $Q=\{q_{t,j}\}$ similarly randomly sampled from the video.

\paragraph{Architecture.}
For a given track $j$, we first embed all locations $(x_{t,j},y_{t,j},t)$ with a sinusoidal embedding across space and time and project to $C$ channels.  Then we add a ``readout'' token of length $C$, and perform self-attention to all tokens for track $j$, using $(1-o_{t,j})$ as an attention mask, which helps the representation be invariant to occluded points. After self attention, we discard all tokens except the ``readout'' token to obtain a fixed-length $C$-channel representation of each track. Next, we use a Perceiver~\citep{jaegle2021perceiver}-style approach to encode all track tokens, namely by cross-attending from a learned set of 128 latent tokens to all track readout tokens, followed by self-attention.  Finally, we project the latent tokens to a lower dimension resulting in a fixed-size $128 \times 64$ dimensional representation $\phi_S$ of the tracks.

To ensure this representation encodes the dense motion, we train it to reconstruct tracks from the video using a decoder. However, since we want to represent the motion as a dense field, we want it to be \textit{invariant} to the specific query points used to obtain the support tracks. To achieve this, we let the decoder reconstruct held-out tracks not included in the input.  That is, the decoder takes $\phi_S$ and a query point $(x_q,y_q,t_q)$, and outputs the track that goes through this point.  We up-project the tokens in $\phi_S$ to a higher dimension with an operator $U$, and compute a ``readout'' token via a sinusoidal position encoding of $(x_q,y_q,t_q)$.
Next, we apply self-attention to the full token set, and discard all but the readout token. 
We finally apply a linear projection to the readout token to obtain $(x^q_t,y^q_t,o^q_t)$ for every frame, where $o^q_t$ is an occlusion logit trained by sigmoid cross-entropy, and $x^q_t,y^q_t$ are trained with an $L_1$ loss.
In Appendix \ref{supp:trajan}, we provide more details about training and architecture.

\paragraph{Average Jaccard.}
An advantage of \model is that it not only computes representations that describe the motion contained in a video, but that it can also be used to evaluate the motion directly: the model was trained on real motions, so we expect unrealistic motions to be more difficult to reconstruct.
Thus, we propose Average Jaccard~\citep{doersch2022tap} (AJ) to measure the accuracy of the autoencoder reconstructions relative to the tracks it receives as input.  Average Jaccard combines occlusion and position accuracy.  For a given threshold $\delta$ $\mbox{Jaccard}_{\delta}$ considers ``true positives`` ($TP$) to be predictions which are within $\delta$ of the ground truth.  ``False positives`` ($FP$) are predictions that are farther than $\delta$ from the ground truth (or the ground truth is occluded), and ``false negatives`` ($FN$) are ground truth points where the prediction is farther than $\delta$ (or occluded).  $\mbox{Jaccard}_{\delta}$ is $TP/(TP+FP+FN)$, and Average Jaccard averages $\mbox{Jaccard}_{\delta}$ over several thresholds (see also Appendix \ref{supp:trajan}).\looseness=-1

\subsection{Alternative models}
\label{sec:alternatives}

\begin{figure*}[btp]
    \captionsetup[subfigure]{justification=centering}
    \centering
    \begin{subfigure}[t]{0.18\textwidth}
        \centering
        \includegraphics[height=1.0in]{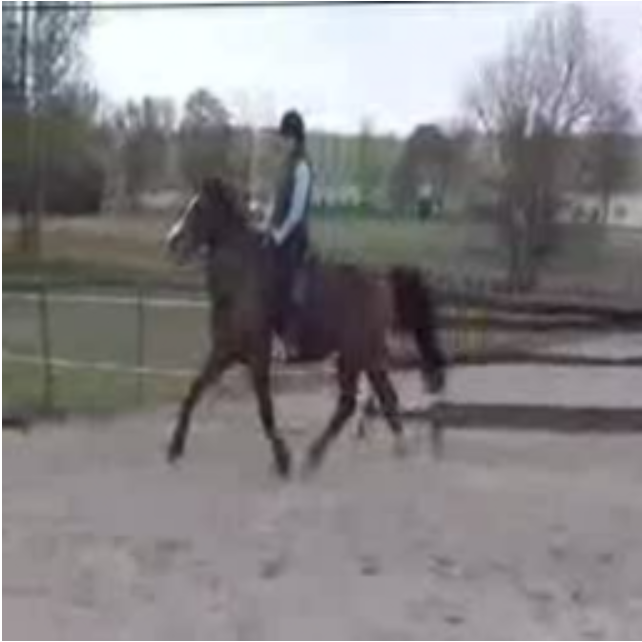}
        \caption{Level 1.1}
    \end{subfigure}%
    ~ 
    \begin{subfigure}[t]{0.18\textwidth}
        \centering
        \includegraphics[height=1.0in]{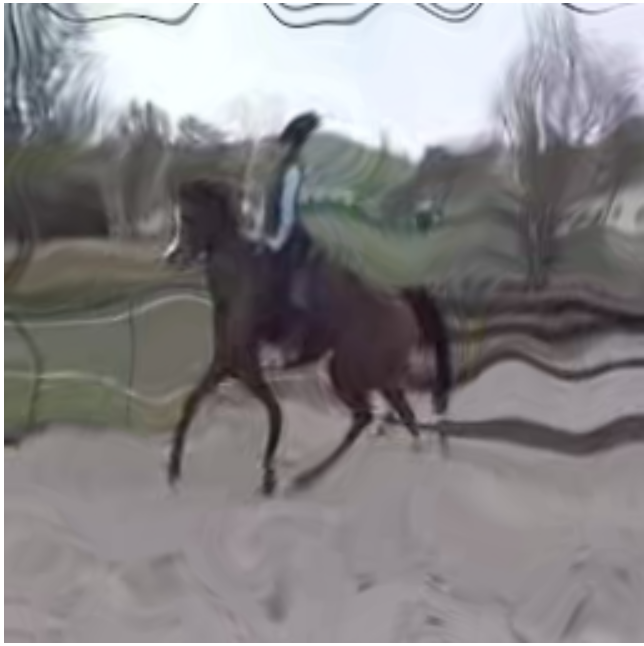}
        \caption{Level 1.2}
    \end{subfigure}
    ~        \begin{subfigure}[t]{0.18\textwidth}
        \centering
        \includegraphics[height=1.0in]{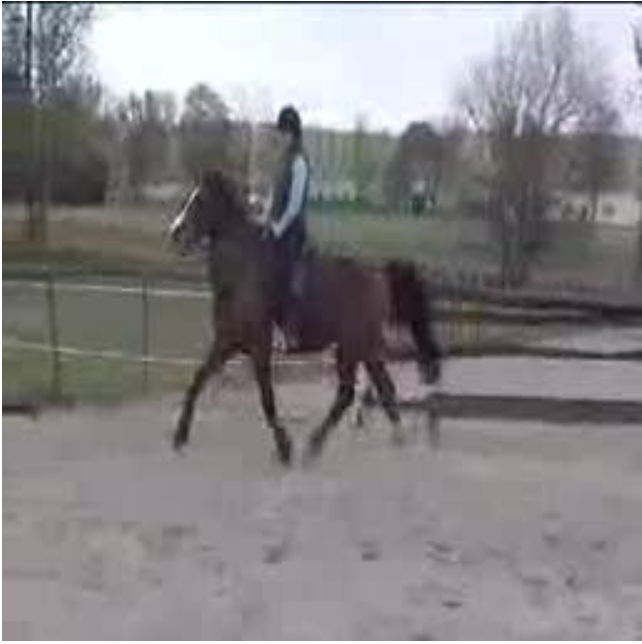}
        \caption{Level 2.1}
        
        \end{subfigure}
                \begin{subfigure}[t]{0.18\textwidth}
        \centering
        \includegraphics[height=1.0in]{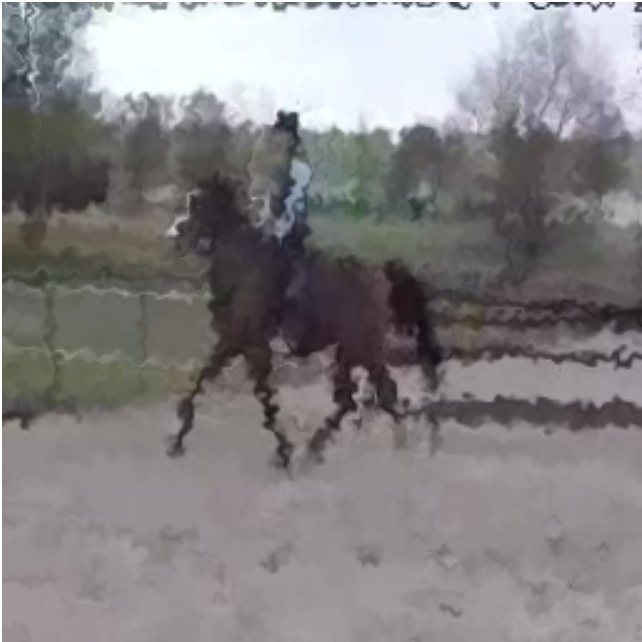}
        \caption{Level 2.2}
        
        \end{subfigure}
        \begin{subfigure}[t]{0.18\textwidth}
        \centering
        \includegraphics[height=1.0in]{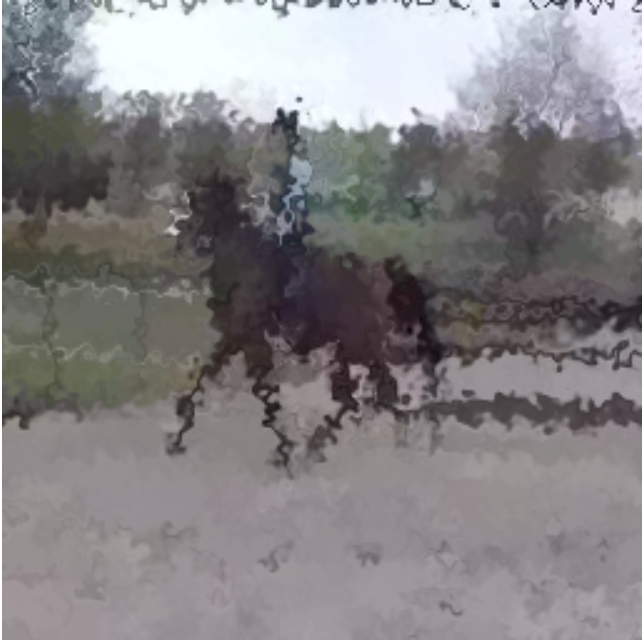}
        \caption{Level 2.3}
        
    \end{subfigure}
    \caption{Synthetic corruptions used to assess sensitivity of evaluation metrics to motion, following \citet{ge2024content} for the UCF-101 dataset.
}
    \label{fig:corruptions}
\end{figure*}
\begin{figure*}[t]
    \centering
    \includegraphics[width=\linewidth]{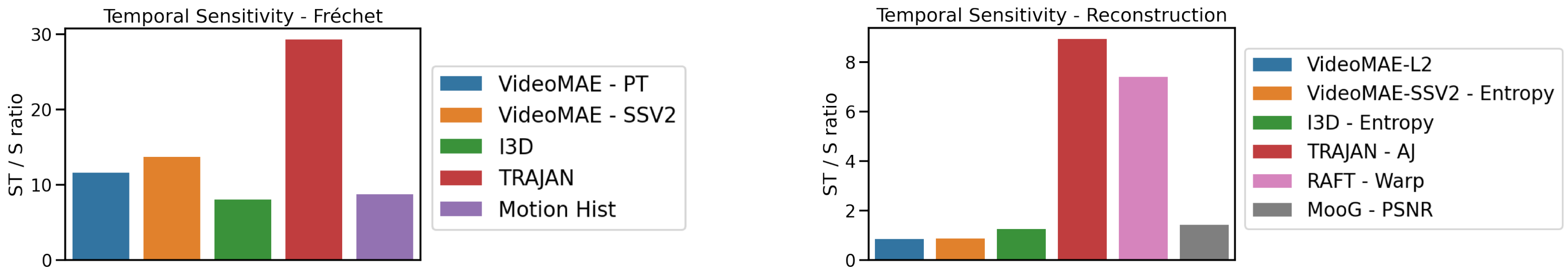}
    \caption{Comparing different methods for detecting temporal distortions on the UCF-101 dataset \citep{ge2024content} in terms of (left) Fr\'echet distances in latent space and (right) per-video ordinal scores. \model and RAFT warp error are particularly sensitive to temporal distortions, while motion histograms~\citep{liu2024frechet} and appearance-based methods perform worse.}
    \label{fig:ucf101}
\end{figure*}

\subsubsection{Motion-based}
For comparison to \model, we consider two alternative motion-based metrics: one based on \emph{histograms} over point track trajectories, and one based on optical flow.

\paragraph{Motion histograms}
\citet{liu2024frechet} propose to evaluate motion consistency in generated videos by estimating velocity and acceleration from point tracks.
Inspired by HOG features~\citep{dalal2005histograms}, they partition the resulting volumes into $4\times5\times5$-sized tubelets and accumulate the magnitude of the values at each angle (using 8 bins) within a tubelet.
Motion features are obtained by concatenating the resulting 1D histograms obtained for each tubelet using both velocity and acceleration. 
We apply this approach to 16-frame chunks of $64\times64$ densely sampled point tracks obtained from BootsTAPIR~\citep{doersch2024bootstap} to yield a 9216-dimensional vector describing the motion within the corresponding 16-frame video.

\paragraph{Optical flow}
Another natural candidate for evaluating motion in videos is to make use of optical flow. 
Unlike point tracks, flow is usually estimated between consecutive frames, which might affect its ability to consider long-term motion patterns. 
Prior work has examined the \emph{warping error} as a loss to make videos more temporally consistent \citep{lai2018learning}, and was used more recently as a metric of temporal consistency in generated videos~\citep{liu2024evalcrafter}.

For a given pair of frames, the warping error is obtained by computing the pixel-wise difference between the second frame and the ``warped prediction'' computed using the optical flow prediction applied to the first frame, while also accounting for predicted points of occlusion.
Following \citet{liu2024evalcrafter}, we calculate the warp differences on every two frames, and average across pairs to obtain the final score. 
To estimate optical flow we use a version of RAFT~\citep{teed2020raft} with added improvements~\citep{sun2022disentangling, saxena2024surprising} (see Appendix \ref{app:experimental-details} for details).
RAFT iteratively updates a flow field based on multi-scale 4D correlation volumes computed from learned features for all pairs of pixels.
Due to the size of these volumes, we only consider the negative warping error (such that higher error corresponds to a lower ordinal value) as a per-video metric as calculating a covariance matrix would otherwise be intractable.\looseness=-1

\subsubsection{Appearance-based}

Inspired by FVD, we also consider a number of methods that operate directly on RGB.
Although such methods are freely able to focus on content or motion related information, prior work encountered a content-bias~\citep{ge2024content}, especially for classification models.

\paragraph{I3D}

The I3D network is an inflated 3D convolutional neural network based on the Inception architecture for image classification that can be applied to videos~\citep{carreira2017quo}.
\citet{unterthiner2018towards} proposed to use the logits of an I3D network that was trained to perform action recognition on the Kinetics-400 dataset (YouTube videos of humans performing various actions~\citep{carreira2017quo}) as a feature space for comparing videos. 
For a single-video metric from I3D, we use the negative entropy over the action class predictions (meaning lower scores indicate worse quality).

\paragraph{VideoMAE}
VideoMAE~\citep{tong2022videomae} is a Masked Autoencoder (MAE) trained on videos using a self-supervised reconstruction objective, which was previously found to reduce frame-level bias relative to I3D when used for evaluating videos \citep{ge2024content}. Here we use VideoMAE-v2~\citep{wang2023videomae}, which is trained on a mixed set of raw video datasets. Importantly, the reconstruction objective for VideoMAE-v2 is not autoregressive -- it is not trained to predict the future, rather it is trained to ``fill in'' missing patch tubes from a video. In practice, VideoMAE-v2 is usually further fine-tuned on specific downstream tasks to deal with domain shift.

We consider two variants of VideoMAE-v2, the pre-trained model (VideoMAE$_{PT}$), and a model that has been further fine-tuned for action recognition on the SSv2 dataset \citep{goyal2017something} (VideoMAE-SSV2). For VideoMAE-PT, we pool over patches to calculate the embeddings, and use the negative $L_2$ loss on the decoded patches as the single-video metric. For VideoMAE-SSV2, we follow \citet{ge2024content} and choose the penultimate layer before classification for the embeddings, and the negative entropy over action class predictions as the single-video metric.

\paragraph{MooG}
MooG~\citep{vanSteenkiste2024moving} is a recurrent model trained for next-frame prediction. It operates by maintaining and updating an internal state constituting of off-the-grid latent tokens, which can be decoded to predict the next frame. These latents are first randomly intialized, and then updated on each iteration by a transformer model which cross attends to the image features of the corresponding frame, followed by a set of self-attention layers. The decoder converts the latent state back to pixels by querying the latents through cross-attention with fixed grid-based features~\citep{sajjadi2022scene,jaegle2022perceiver}.
We slightly modify the original model by decoding the next-frame from the ``corrected'' state to simplify the implementation. 
\citet{vanSteenkiste2024moving} showed how MooG learns representations that encode both appearance and motion, as they can be decoded into both reconstructed video and point-tracks using shallow decoder heads.
However, due to the size of the latent space, we will only consider its prediction error in the form of PSNR as a single-video metric.

\section{Human evaluation}
\label{sec:method-humans}

To evaluate whether the proposed metrics are useful, we compare them to human judgements.
We propose to source fine-grained human annotations for generated videos, focusing on the following dimensions.
\emph{Consistency:} Whether the appearance and motion of objects and background appear consistent over time;
\emph{Interactions:} Whether the interactions between the objects are realistic (if any take place);
\emph{Realism:} Whether the objects and background are depicted realistically; and
\emph{Speed:} Whether the motion of the objects and camera is perceived as being slow, normal, or fast. 

Together, these are intended to capture common failure modes of generative video models, such as objects changing in shape unnaturally, discrete jumps or jitter, poor interactions, such as objects that blend when they come together, or highly implausible scenes.
We allow humans to provide real-valued judgements where possible to reflect that a generated video might typically meet the criteria above only to a variable degree.
We refer to Appendix~\ref{app:experimental-details-human-study} for details about questions, including an overview of the evaluation UI.\looseness=-1

%% file: sections/5_results.tex
\begin{figure*}[t]
    \centering
    \begin{subfigure}[t]{0.48\textwidth}
        \centering
        \includegraphics[height=1.20in]{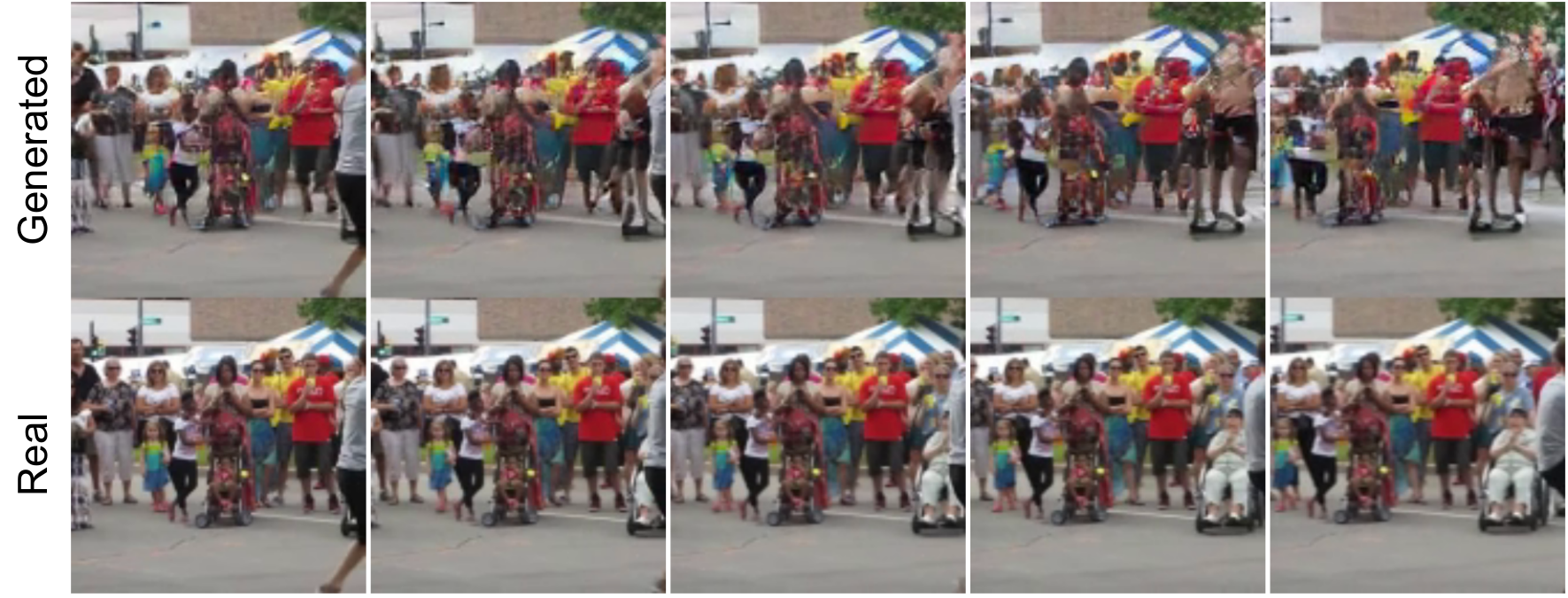}
        \caption{A generated video from WALT where the camera pans to new parts of the scene that were not initially visible (e.g.\ the wheelchair). The objects on the right are inherently impossible to predict, which leads to a large pixel error between the generated and the real video. However, since the \emph{camera motion} is correct, the distance between their \model embeddings is small.}
    \end{subfigure}
    ~
    \begin{subfigure}[t]{0.48\textwidth}
    \centering
        \includegraphics[height=1.20in]{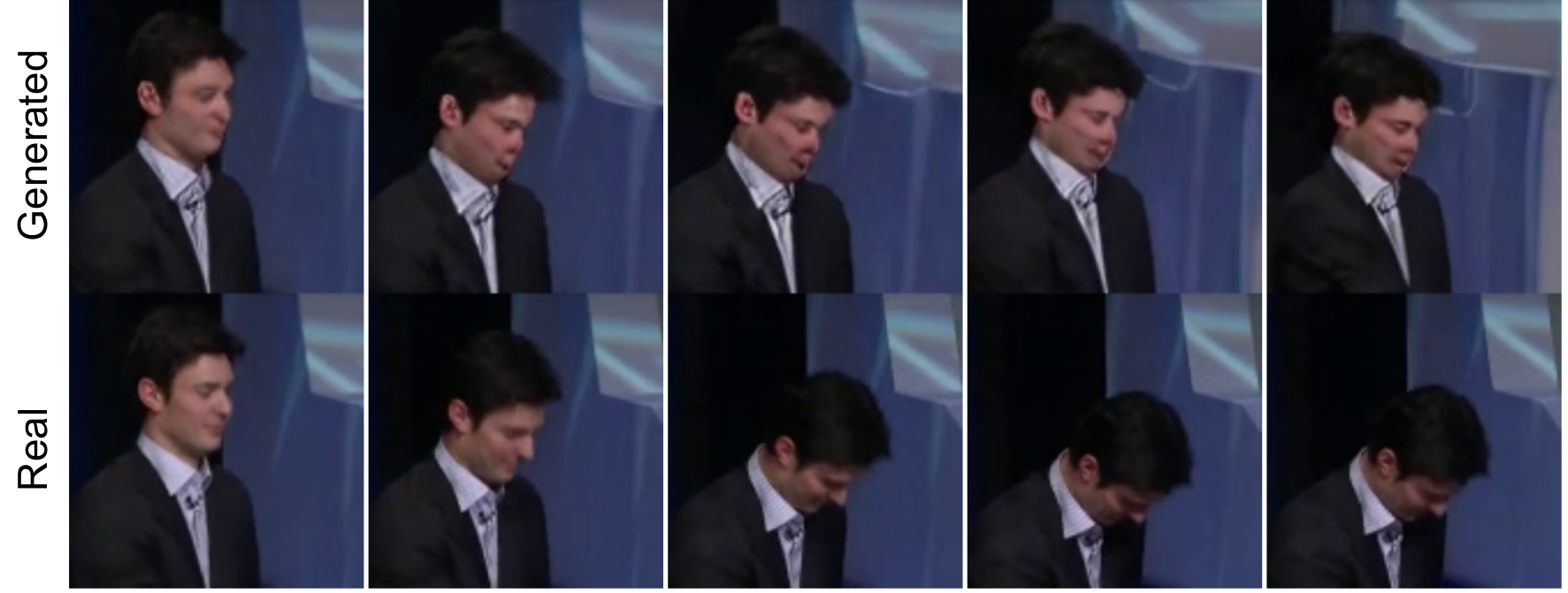}
        \caption{A generated video from WALT with incorrect predicted motion. Despite the incorrect motion, the I3D and VideoMAE latent space distance between the generated and real videos is very low (rank 38, 74 / 2364), while the distance in the \model embeddings is high (rank 1655 / 2364).}
    \end{subfigure}
    \captionsetup{justification=centering}
        \caption{WALT extrapolations for two videos (top: generated, bottom: real).}
    \label{fig:walt-compare}
\end{figure*}

\section{Results}

We evaluate \model and alternative approaches in three different scenarios for how one might make use of them as metrics in practice: comparing videos at a distributional-level (\S\ref{sec:results-distributional}), comparing generated to real videos (\S\ref{sec:video-to-video}), and per-video (\S\ref{sec:results-per-video}) quality assessments.
Across all settings, we find that \model consistently performs better than the alternatives in its sensitivity to motion artefacts.
We report additional results in Appendix~\ref{app:additional-results} and complete experimental details in Appendix~\ref{app:experimental-details}.

\subsection{Comparing video distributions}
\label{sec:results-distributional}

We start by comparing empirical distributions of videos, which is the usual paradigm for metrics based on FVD~\citep{unterthiner2018towards}. 
We obtain real videos from the UCF101 dataset, which consists of 13,320 videos recorded in the wild that show humans performing different types of actions~\citep{soomro2012ucf101}, and synthetically corrupt them using the 5 levels of ``elastic transformations'' from \citet{ge2024content}.
Corruption levels 1.1 and 1.2 apply low frequency deformations, while 2.1, 2.2, and 2.3 apply high frequency deformations (Figure \ref{fig:corruptions})\footnote{Parameters were obtained via personal correspondence.}.
There are two corruption modes: a spatiotemporal mode, where video frames are distorted independently using different parameters, and a spatial mode where the same set of parameters is used to distort all the frames.
\citet{ge2024content} propose to use the \emph{ratio} of these modes to measure how sensitive metrics are to temporal corruptions.

We first apply the standard technique of computing the Fr\'echet distance between the latent representations computed for videos within each set, using \model and four of the models presented in \S\ref{sec:alternatives}. 
\autoref{fig:ucf101} (left) reports the average temporal sensitivity across different levels of corruption strength as in \citet{ge2024content}.
TRAJAN is the most sensitive to temporal distortions.

\begin{figure*}
    \centering

        \includegraphics[height=1.1in]{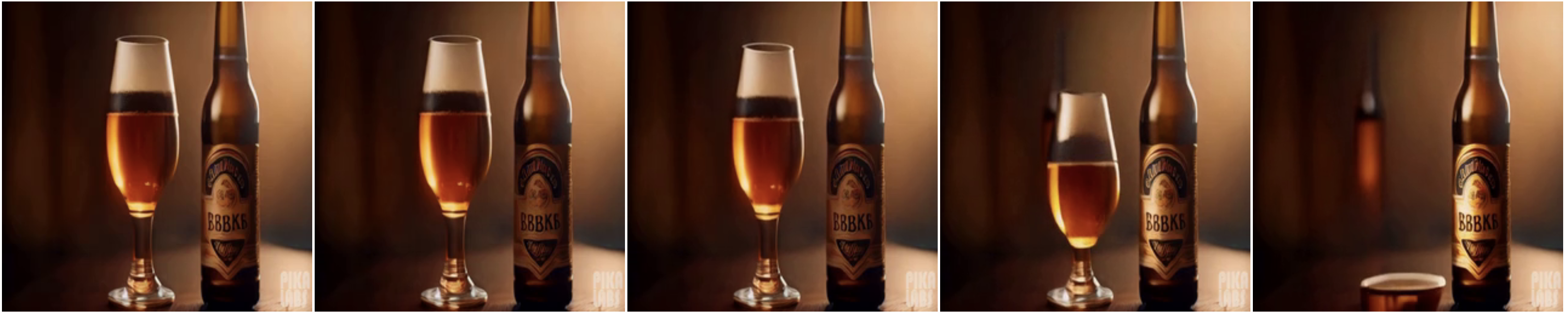}
        \caption{Humans rate this video to be poor in terms of all consistency / realism questions. All automated metrics score this positively. In this case, none of the metrics are capturing the unexpected disappearance and morphing of the glass.}
    \label{fig:failures}
    \end{figure*}
To bridge distribution level and single-video metrics, we compute the latter for each video and take the average across the set (\autoref{fig:ucf101}, right).
TRAJAN and the RAFT warp error are particularly sensitive to temporal distortions, while motion histograms~\citep{liu2024frechet} and appearance-based methods are not.
Although distributional comparisons yield the highest overall sensitivity, the strong performance of \model suggests that per-video metrics based on motion can be used as a replacement for distribution-based metrics such as FVD if a reference dataset is not available or if sample sizes are limited.

\subsection{Comparing real and generated video pairs}
\label{sec:video-to-video}

\begin{table}
    \small
    \centering
    \begin{tabular}{ccc}
    \toprule
    Method & PSNR & SSIM \\
    \midrule
       V-MAE$_{PT}$  & 0.35 & 0.53 \\
       I3D & \textbf{0.24} & 0.26 \\
       TRAJAN & \textbf{0.23} & \textbf{0.13} \\
       \bottomrule
    \end{tabular}
    \caption{ Spearman's $\rho$ between latent space distances and PSNR / SSIM scores for high motion videos generated by WALT. }
    \label{tab:vid-to-vid-corr}
\end{table}

Many video models are evaluated by conditioning on a few frames of a \emph{real} video, and then predicting how that video should unfold in the future \citep{whitney2023learning, wang2017predrnn, lin2020improving}. However, comparing future generated and real frames is not straightforward. Comparisons in pixel space can suffer from inherently uncertain futures such as objects that might enter the frame, or panning a camera such that new parts of the scene are visible (\autoref{fig:walt-compare}a). In this section, we investigate whether comparisons in the latent space of various models can capture video-to-video comparisons of realism better than pixel-based scores.

\begin{table*}[t]
\small
\centering
\begin{tabular}{lcccccccc}
\toprule
 & \multicolumn{4}{c}{EvalCrafter} & \multicolumn{4}{c}{VideoPhy} \\
 \cmidrule(lr){2-5} \cmidrule(lr){6-9}
 \multirow{2}{*}{Method}  & Motion & App. & Realism & Interacts & Motion & App. & Realism & Interacts \\
 & Consist. & Consist. &  & & Consist. & Consist. & & \\
\midrule
V-MAE$_{PT}$ - $L_2$ & -0.00 & 0.00 & 0.04 &-0.05 &  -0.03 & -0.06 & 0.06 & 0.02\\
I3D - Entropy & -0.01 & -0.03 & -0.02 & 0.03 &  0.09 & 0.08 & 0.14 & \textbf{0.09}\\
RAFT - Warp & \textbf{0.28} & 0.27 & 0.25 & 0.13 & 0.20 & 0.26 & 0.18 & 0.03\\
MooG - PSNR & 0.21 & 0.19 & 0.17 & 0.05 & 0.11 & 0.16 & 0.07 & 0.01 \\
TRAJAN & \textbf{0.29} &\textbf{0.29} & \textbf{0.27} & \textbf{0.19} & \textbf{0.25} & \textbf{0.32} & \textbf{0.29} & \textbf{0.09}\\
\midrule
Inter-rater $\sigma$ & 0.49& 0.46& 0.47& 0.53 & 0.48 & 0.46 & 0.48 & 0.53 \\
\bottomrule
\end{tabular}
\caption{Spearman's rank coefficients between human ratings and automated metrics for a subset of videos from EvalCrafter \citep{liu2024evalcrafter} and VideoPhy \citep{bansal2024videophy} (higher is better). Inter-rater $\sigma$ is the standard deviation of human responses (lower is better).}
\label{tab:human-consistency-realism}
\end{table*}

To do so, we take the WALT video diffusion model \citep{gupta2025photorealistic} and train it on the Kinetics-600 dataset~\citep{carreira2018short} to predict future video frames conditioned on 2 real latent frames. We sample checkpoints from throughout training, and calculate video embeddings using \model, VideoMAE and I3D on 2364 generated and corresponding real samples with high motion as estimated by the length of the point trajectories.

In \autoref{tab:vid-to-vid-corr}, we show that the distances in latent space are not well correlated with either PSNR or SSIM for any of the models, and the correlation is lowest for \model. This suggests that the \model latent space is capturing something fundamentally different from pixel error, which we highlight with examples in \autoref{fig:walt-compare}. 
In particular, \model is sensitive to differences in \emph{motion} but not appearance, while distances in the VideoMAE and I3D latent spaces are more sensitive to overall recognizability.
This suggests that the \model latent space is useful for measuring similarities in the motion between videos, even when the exact pixels or overall appearance are not a perfect match.

\subsection{Assessing generated videos individually}
\label{sec:results-per-video}

In many cases, we want a metric that can be used to determine the quality of the motion for a single video, i.e. without a reference.
It is increasingly common to encounter generative video models for which we do not have access to training data, or that are too computationally expensive to sample many videos from.
Here we investigate metrics to evaluate different aspects of motion on a per-video basis.
We source generated videos from the ``solid-solid'' split of VideoPhy~\citep{bansal2024videophy}, which contains videos generated by 8 different models, and EvalCrafter~\citep{liu2024evalcrafter}, using 11 different models.
Although these datasets include human evaluations, they do not contain enough information to evaluate consistency between raters, and describe general motion-related dimensions in aggregate.
Therefore, in addition to the original ratings, we conduct our own human evaluation focusing on individual motion dimensions as outlined in \S\ref{sec:method-humans}.
Each video is rated by 3 raters sampled randomly from a total pool of 10.
We then z-score each rater's responses to account for different raters using the sliding scale differently (Appendix \ref{app:experimental-details-human-study}). 

\begin{table}[h]
\centering
\begin{footnotesize}
\begin{tabular}{lcccc}
\toprule
 & \multicolumn{2}{c}{EvalCrafter} & \multicolumn{2}{c}{VideoPhy} \\
 \cmidrule(lr){2-3} \cmidrule(lr){4-5}
 & Camera & Object & Camera & Object \\
 Method & speed & speed &  speed & speed  \\
\midrule
RAFT - Mag. & 0.32 & \textbf{0.57} & 0.12 & \textbf{0.63} \\
\model - Len. & 0.26 & \textbf{0.58} & 0.16 & 0.47 \\
\model - Radii & \textbf{0.42} & 0.01 & \textbf{0.39} & -0.28 \\
\midrule
Inter-rater $\sigma$ & 0.06 & 0.08 & 0.07 & 0.08 \\
\bottomrule
\end{tabular}
\end{footnotesize}
\caption{Spearman's rank coefficients between human ratings and automated metrics for the \emph{amount of motion} in generated videos from the EvalCrafter and VideoPhy datasets.}
\label{tab:human-speed}
\end{table}

\begin{figure*}
    \centering
    \includegraphics[width=0.8\textwidth]{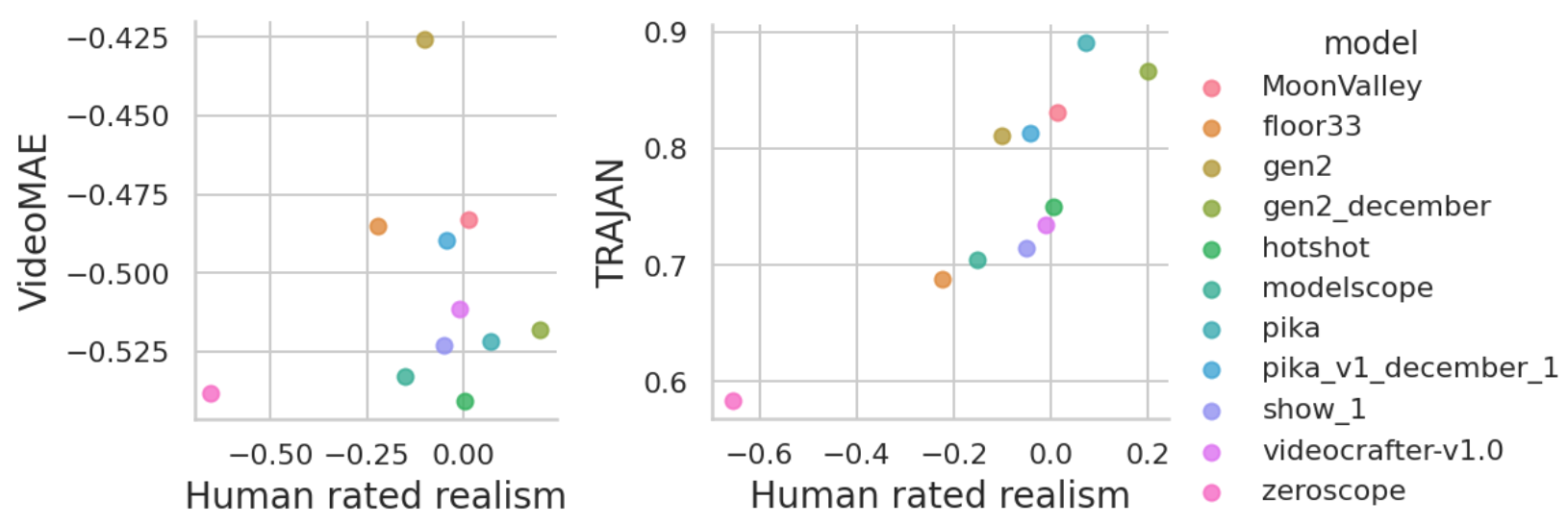}
    \caption{TRAJAN (right) captures the rank ordering of human preferences for different generative video models very well relative to VideoMAE (left). When correlated with human realism ratings, TRAJAN achieves a spearman rank coefficient of 0.9, while VideoMAE only achieves 0.07.}
    \label{fig:overallmodels}
\end{figure*}

A summary of our main findings are shown in \autoref{tab:human-consistency-realism} for consistency related dimensions, and \autoref{tab:human-speed} for speed related dimensions.
We also show overall correlations for \model and VideoMAE against human-rated realism for the 11 different generative video models from EvalCrafter in \autoref{fig:overallmodels}.
All numbers are reported as Spearman's rank coefficients.
We find interesting and surprising results from the human study.
First, the humans themselves are not consistent with each other.
We do not have sufficient numbers of overlapping examples for rater pairs, so we instead compute the inter-rater standard deviation $\sigma$ for each video for each question type and average across all videos in the dataset.
Larger $\sigma$ therefore indicates more discrepancy between raters for a given question type.
For example, given that ratings were z-scored by participant, the standard deviation for a single rater \emph{across the dataset} would be 1.0. A $\sigma$ of 0.5 therefore indicates that there is approximately half as much variation in rater responses to a single question as there is variation in a single rater across the dataset.

Across all question types, in both datasets, TRAJAN best correlates with human judgements, and performs similar to RAFT for consistency-related categories.
While this may be expected for motion and appearance consistency, \model also correlates best for \emph{realism} both for the video overall, and for the object interactions.
This holds not just on a per-video basis, but also tracks which generative video model humans prefer overall (\autoref{fig:overallmodels}).
If we condition on only high motion videos (those rated by humans as having camera speeds greater than or equal to ``medium''), where we expect motion-based metrics to be most informative, we see even stronger correlations (and lower inter-rater $\sigma$s), especially for the EvalCrafter dataset (\autoref{evalcrafter-controlled} and Appendix~\ref{app:additional-results}).
The point tracks can also be used to measure motion \emph{amount} (\autoref{tab:human-speed}), with both point track lengths and optical flow magnitude being predictive of object speeds, and point track radii (see Appendix~\ref{supp:trajan} for details on how these are calculated) being predictive of camera speed.

However, even TRAJAN is not a perfect predictor of human judgements.
For example, in \autoref{fig:failures}, a beer glass disappears by collapsing into the table. All metrics rate this video highly, as the motion is smooth and easily tracked. However, this is a very unrealistic motion, since there is nothing that would cause the glass to collapse.

\begin{table}[]
    \footnotesize
    \centering
    \begin{tabular}{cccccc}
    \toprule
    Method     &  Consist. & Quality & Visual & T2V & Subj. \\
    \midrule
V-MAE$_{PT}$ & -0.19 & -0.16 & -0.06 & -0.07 & -0.02 \\
I3D & 0.09 & \textbf{0.24} & 0.22 & \textbf{0.25} & 0.14 \\
RAFT & 0.08 & -0.01 & 0.09 & 0.13 & 0.15 \\
MooG & 0.13 & 0.08 & 0.01 & 0.11 & 0.03 \\
TRAJAN & \textbf{0.33} & \textbf{0.24} & \textbf{0.28} & \textbf{0.25} & \textbf{0.23} \\
    \bottomrule
    \end{tabular}
    \caption{Spearman rank coefficients on the original EvalCrafter dataset for their categories of temporal consistency, motion quality, visual quality, text-to-video similarity, and subjective likeness, when holding the overall amount of motion fixed.\vspace{-1.5em}}
    \label{evalcrafter-controlled}
\end{table}

\paragraph{Original EvalCrafter dataset} To ensure that our metrics are not biased to the human evaluations we collected, we also compare to the human ratings released on the EvalCrafter dataset by the original authors in Appendix~\ref{app:additional-results}. However, we found that human judgements in this dataset (across temporal consistency, visual quality, text-to-video similarity, and subjective likeness), were mostly driven by simple measures of overall motion (i.e.\ videos with significant motion are worse, regardless of how realistic that motion appears). We therefore re-analyze the data controlling for overall motion by binning videos based on their optical flow magnitude (RAFT - Mag), and then evaluate Spearman rank coefficients within bins falling into the top half of the dataset (those where there was significant motion). We average the coefficients across bins. Here, TRAJAN strongly outperforms other metrics in explaining temporal and visual consistency, as well as subjective likeness, suggesting that it captures more than just overall motion magnitudes in explaining motion quality.

\begin{figure}
    \centering
    \includegraphics[width=0.9\linewidth]{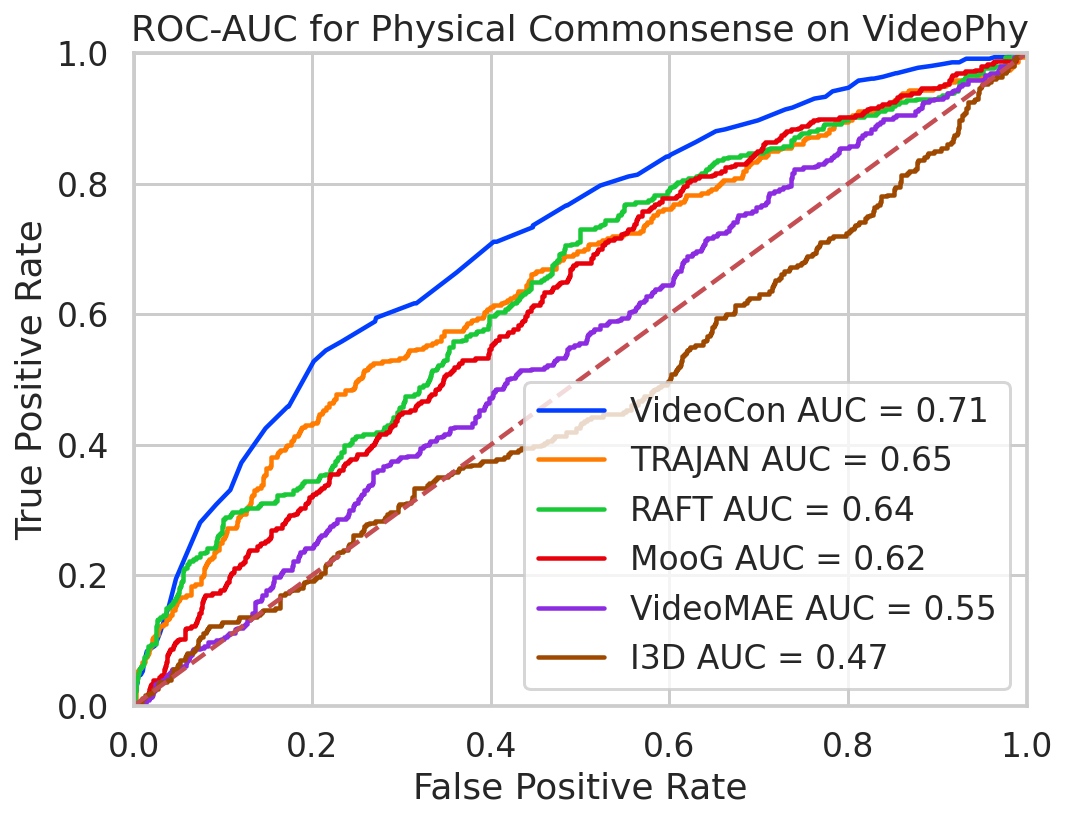}
    \caption{ROC-AUC Scores for physical consistency on VideoPhy dataset. Note that VideoCon is \emph{fine-tuned} on human labels, whereas all other methods are not. \vspace{-1.5em}}
    \label{fig:rocaucvideophy}
\end{figure}

\paragraph{Original VideoPhy dataset} Finally, as a further external validation, we report the area under the ROC curve using the original human labels for physical consistency contained in the VideoPhy dataset (\autoref{fig:rocaucvideophy}).
RAFT, MooG and \model are the top-performing models in this case.
We also compare to VideoCon as proposed in \citet{bansal2024videophy} as a method for evaluating physical consistency in generated videos, which fine-tunes a vision-language model on a subset of the human labels.
Without any fine-tuning on human data, \model performs strikingly well, which indicates its generality for evaluating motion in videos originating from different models and datasets (i.e. prompts).

\begin{figure*}[t]
    \centering
    \includegraphics[width=\textwidth]{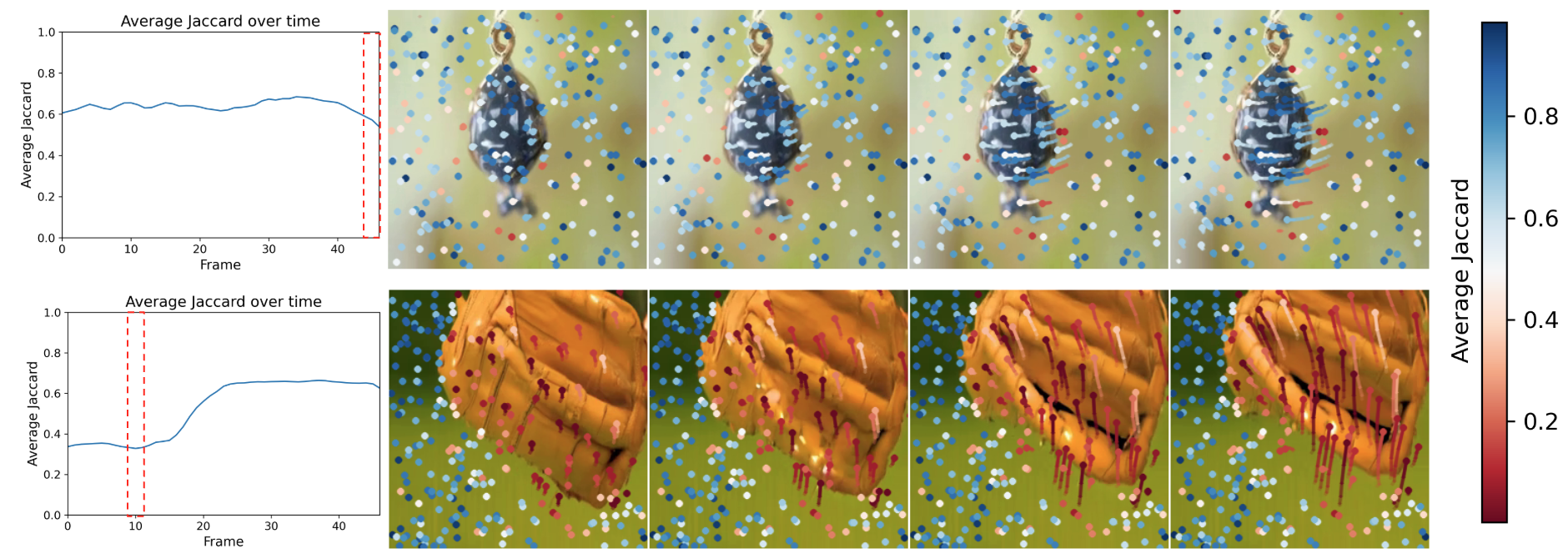}
    \caption{The Average Jaccard (reconstruction quality -- higher AJ means higher reconstruction quality) from TRAJAN can be used to detect generated video inconsistencies at specific points in space and at specific moments in time. Here we show two examples (top and bottom). On the \textbf{left} the AJ is averaged over all points for each frame of the video. This  highlights moments in time where there are temporal inconsistencies overall. On the \textbf{right} we visualize the AJ for each visible point track for 4 consecutive frames centered on the frame having the lowest AJ score (indicated with the dashed lines on the left). Red indicates poor reconstruction (low AJ) while blue indicates good reconstruction (high AJ). In the top case, the motion is consistent over most of space, so there is overall higher AJ across the points. In the bottom case, the glove's appearance morphs over the frames, leading to worse reconstruction for points located on and immediately around the glove.}
    \label{fig:detailed}
\end{figure*}
\subsection{Spatiotemporal error localization with TRAJAN}
Besides overall correlations between TRAJAN reconstruction quality and human participants, we can also use TRAJAN to localize generative video errors in both space and time. In particular, TRAJAN allows us to calculate reconstruction quality (through the Average Jaccard metric applied to the reconstructed points) for \textit{each point in space} over \textit{short windows in time} to determine \textit{where} and \textit{when} motion and appearance inconsistencies occur.

We demonstrate this in \autoref{fig:detailed} for a generated video which TRAJAN scores as being highly consistent (top) and highly inconsistent (bottom). Looking at the progression over time on the left, the consistent video maintains reasonably high Average Jaccard. However, for the video at the bottom, where the glove's appearance morphs throughout the beginning of the video before stabilizing, we see a dramatic change in Average Jaccard. Visualizing the spatial errors around these poorly reconstructed frames (on the right) shows that the points on the glove are contributing the most to the low score.

%% file: sections/6_conclusion.tex
\section{Conclusion}
We investigated different methods for evaluating motion in generated videos, and proposed \model, a novel architecture for auto-encoding dense point tracks.
We showed that \model is more than three times more sensitive to synthetic temporal distortions when used with the Fr\'echet distance, that it can be used to compare motions of generated and real video pairs even when the appearance is not maintained, and that its reconstruction error correlates remarkably well with human judgements of motion consistency, appearance consistency, realism, and even visual quality.
Together, these results suggest an alternative approach to evaluating generated videos: not just by using a different backbone for FVD, but by moving away from costly distributional comparisons all together, and focusing on comparing generated / real video pairs in latent space or evaluating individual videos (all three are supported by \model).\looseness=-1

Despite the strengths of \model, we also found that there are shortcomings.
By conducting a detailed human study, we showed that while \model predicted human judgements better than a range of alternative motion- and appearance-based metrics, there are videos for which no automated metric captures human ratings.
To capture human judgements in these scenarios, future work will need to develop more advanced models that can make more accurate judgements not just about how something \emph{does} move, but how it \emph{should} move if it was obeying physical principles.

Furthermore, individual people seem to care about different aspects of videos for evaluating basic properties like temporal and appearance consistency, leading to low consistency in evaluations across human participants.
This suggests that future work is needed to determine how to elicit better judgements from human raters, and suggests caution for using humans as a gold standard for adjudicating between models.\looseness=-1

\section*{Acknowledgements}

We would like to thank Mike Mozer for helpful comments and suggestions to improve our paper, and Viorica Patraucean, Yiwen Luo, Lily Pagan, and Nishita Shetty for help preparing the human study.
We would like to thank Songwei Ge and Ge Ya Luo for sharing details about the video corruptions used in their work. 

%% file: sections/X_suppl.tex
\section{Additional Results}
\label{app:additional-results}

\subsection{MMD with different backbones}

An alternative to using Fr\'echet distance for comparing embeddings is to make us of the Maximum Mean Discrepancy~\citep{gretton2012kernel} as a distance function.
This idea was first proposed for comparing distributions of images in \citet{binkowski2018demystifying}, and briefly explored for video in \citet{unterthiner2018towards}.
MMD is a kernel-based approach to comparing distributions without assuming a particular form (such as a Gaussianity when using Fr\'echet distance).
An unbiased estimator of the squared MMD is:

\begin{equation}
    \sum_{i \neq j}^{m} \frac{k(x_i, x_j)}{m(m-1)} - 2 \sum_i^m\sum_j^n \frac{k(x_i,y_j)}{mn} + \sum_{i \neq j}^{n} \frac{k(y_i, y_j)}{n(n-1)}.
    \label{eq:mmd}
\end{equation}

where $\{x_i\}_{i}^{m}$ and $\{y_j\}_{j}^{n}$ are samples drawn from the respective distributions we wish to compare, and $k(\cdot,\cdot)$ is a kernel function. Here we use the polynomial kernel function $k(a, b) := (a^Tb+1)^3$ as in  \citet{unterthiner2018towards}.

\begin{figure}[h]
    \centering
    \includegraphics[scale=0.4]{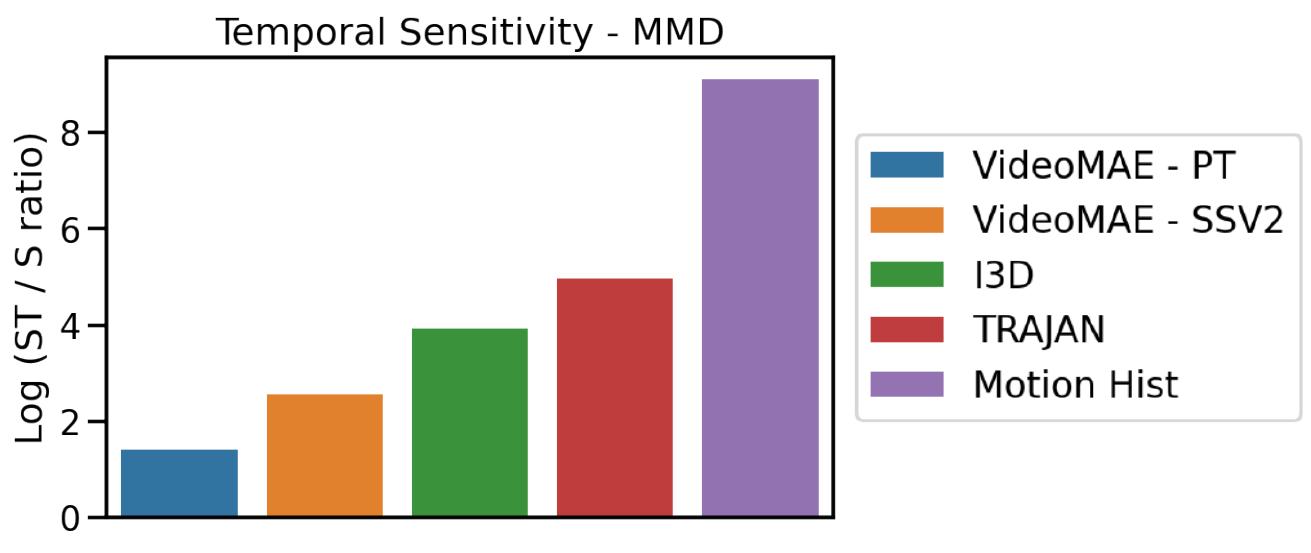}
    \captionsetup{justification=centering}
    \caption{MMD for UCF-101 corruptions}
    \label{fig:mmd-ucf}
\end{figure}

In \autoref{fig:mmd-ucf} we report the results of using MMD instead of the Fr\'echet distance for calculating the sensitivity of different models to temporal corruptions on the UCF-101 dataset.
Defined as in \eqref{eq:mmd}, the MMD can occasionally be negative. If the MMD is negative (which was the case for the motion histograms model for several corruption levels), we replace it with $1e-6$. With MMD, all models are significantly more sensitive to temporal corruptions, with motion histograms showing the largest sensitivity.
Upon further investigation we noticed that this is primarily due to 3 cases where the spatial-only corruptions gave negative MMD scores.

\subsection{Metric changes during training}
\begin{figure}[h]
\captionsetup[subfigure]{justification=centering}
    \centering
    \begin{subfigure}[t]{0.45\textwidth}
        \centering
        \includegraphics[height=1.3in]{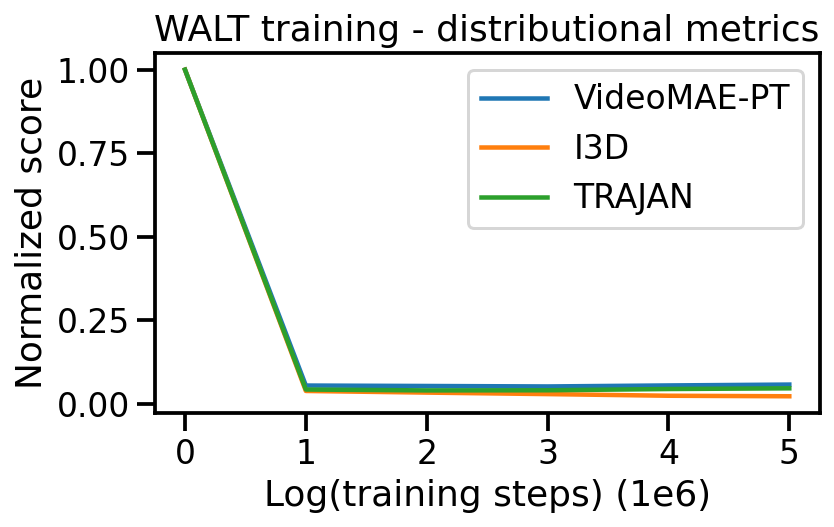}
        \caption{Distribution-level metrics.}
    \end{subfigure}
    \begin{subfigure}[t]{0.45\textwidth}
        \centering
        \includegraphics[height=1.3in]{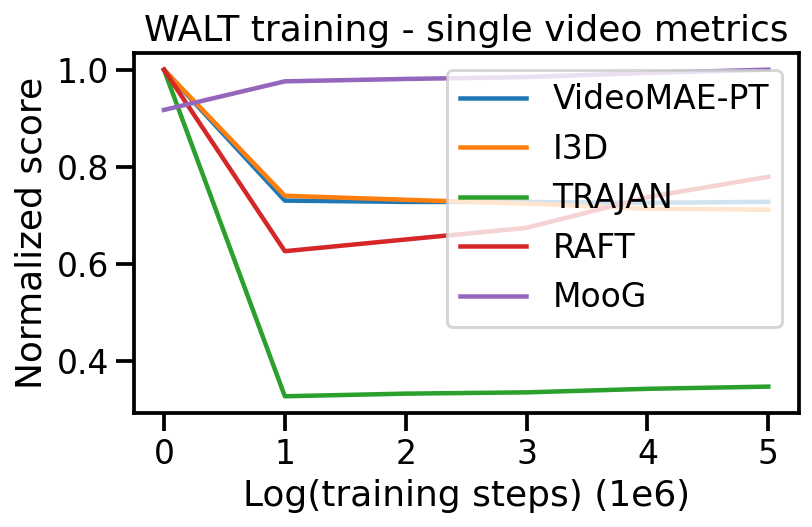}
        \caption{Single-video metrics.}
    \end{subfigure}
\captionsetup{justification=centering}
\caption{Metrics over training iterations for WALT.}
\label{fig:walt_training}
\end{figure}

We measure how each of the different metrics tracks WALT's \citep{gupta2025photorealistic} training progress at 6 different checkpoints in training (\autoref{fig:walt_training}). For the distribution-level metrics, we see similar performance using \model, VideoMAE or I3D. For the single-video metrics, only \model tracks WALT's training reasonably well. However, most of the change in all metrics is driven by the first 100k steps, after which all metrics become more-or-less constant.

\subsection{Additional WALT examples for generated vs. real videos}    
\begin{figure*}[t]
    \centering
    \begin{subfigure}[t]{0.9\textwidth}
    \centering
    \includegraphics[width=0.9\textwidth]{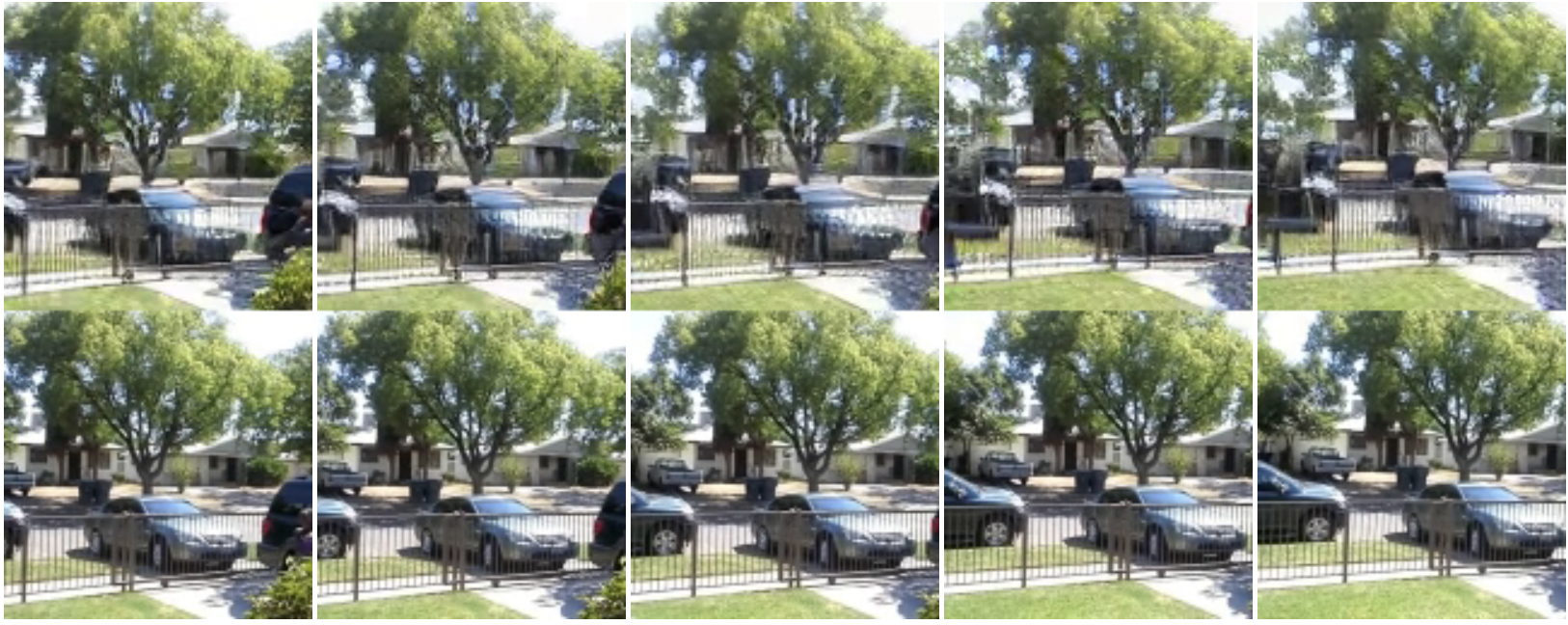}
    \caption{Generated (top) with real (bottom) from WALT. In this example, the camera pans to the left. The pixel error between these two videos is very high, since as the camera moves to the left, it does not fill in the left side of the image successfully, nor does it predict the exact spacing of the grating. However, since the motion of all the objects is correct, the distance between these videos in the \model feature space is low.}
    \end{subfigure}
    \vspace{10pt}
        \begin{subfigure}[t]{0.9\textwidth}
        \centering
    \includegraphics[width=0.9\textwidth]{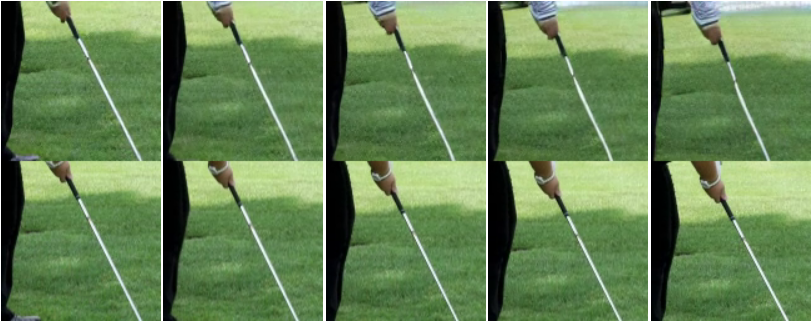}
    \caption{Generated (top) with real (bottom) from WALT. In this example, the golfer's arm moves into the frame from the top. The pixel error between these two videos is high, because the color of the man's jacket is incorrect, and there is a generated background object at the top of the video that does not actually exist. However, the distance in the \model feature space is low, because the motion of the golfer's arm is correct.}
    \end{subfigure}
    \captionsetup{justification=centering}
    \caption{Additional paired WALT generations.}
    \label{fig:additionalwalt}
\end{figure*}

In \autoref{fig:additionalwalt} we show additional examples of comparing real vs. generated videos from the WALT model \citep{gupta2025photorealistic} which have different amounts of average pixel error. In \autoref{fig:additionalwalt}a, there is very high pixel error because of objects that could not be seen as the camera pans to the left, and in \autoref{fig:additionalwalt}b, the pixel error is reasonably high because the predicted color of the golfer's jacket is incorrect, and an extra background object is generated which does not exist. In both cases, the distance between the real and generated video in \model feature space is low, because the motion of the camera (in the first case), or the man (in the second case) is correct.

\subsection{Additional results on VideoPhy \citep{bansal2024videophy} and EvalCrafter \citep{liu2024evalcrafter}}
\autoref{tab:full-evalcrafter} shows the results of different automated metrics on the full EvalCrafter \citep{liu2024evalcrafter} dataset using their labels. We found that across the dataset, RAFT - Mag, the average optical flow magnitude across the video, is a very good predictor of human judgements in all conditions. This suggests that, on average, videos that are faster are more poorly rated by humans generally.
\begin{table}[h]
    \centering
    \small
    \begin{tabular}{c|c|c|c|c}
    \toprule
    Method     &  Consistency & Visual & T2V & Subj. \\
    \midrule
    VideoMAE$_{PT}$ & -0.19 & -0.12 & -0.05 & -0.09 \\
    I3D & -0.00 & 0.01 & 0.00 & -0.01 \\
    RAFT - Warp & \textbf{0.61} & \textbf{0.42} & 0.18 & \textbf{0.34} \\
    RAFT - Mag &  -0.58 & -0.37 & -0.17 & -0.32 \\
    MooG & 0.42 & 0.23 & 0.10 & 0.22 \\
    TRAJAN & 0.52 & 0.36 & \textbf{0.20} & 0.32 \\
    \bottomrule
    \end{tabular}
    \captionsetup{justification=centering}
    \caption{Results on full EvalCrafter \citep{liu2024evalcrafter} dataset.}
    \label{tab:full-evalcrafter}
    \end{table}

\begin{table*}[h]
\centering
\small
\begin{tabular}{l|ccc|ccc}
\toprule
 & \multicolumn{3}{c|}{\textbf{EvalCrafter}} & \multicolumn{3}{c}{\textbf{VideoPhy}} \\
 \midrule
 & \textbf{Motion} & \textbf{Appearance} & \textbf{Realism} & \textbf{Motion} & \textbf{Appearance} & \textbf{Realism} \\
 & \textbf{Consistency} & \textbf{Consistency} &  & \textbf{Consistency} & \textbf{Consistency} &  \\
\midrule
        VideoMAE$_{PT}$ & 0.06  & -0.05 & 0.01 & -0.01 & 0.04 & 0.01\\
        I3D & -0.05 & 0.08 & -0.04 & 0.14& 0.09 & \textbf{0.19} \\
        RAFT & 0.21 & 0.19 & 0.11 & 0.30& \textbf{0.40}&0.09\\
        MooG & 0.10 & 0.08 & -0.01 & 0.19 & 0.31&0.13\\
        TRAJAN & \textbf{0.44} & \textbf{0.51} & \textbf{0.34} & \textbf{0.33} & \textbf{0.41} & 0.14 \\
        \bottomrule
        Rater $\sigma$ & 0.43 & 0.41 & 0.41 & 0.47 & 0.46 & 0.44 \\

\bottomrule
\end{tabular}
\caption{Spearman's rank coefficients between human ratings and automated metrics for the medium - high motion subset of the data. In this subset, people are more consistent in their scores (lower rater $\sigma$), and in most cases, TRAJAN also better predicts human ratings. In the case of VideoPhy realism, there are specific failure modes in the dataset where some models produce videos with inconsistent motion but realistic individual frames, see \autoref{fig:videophy-choppy}, but humans still rate these as realistic.}
\label{tab:human-speed-high}
\end{table*}
\begin{table*}[h]
\small
\centering
\begin{tabular}{lcccc}
\toprule
 Method & Motion Consistency & Appearance Consistency & Realism & Interacts \\
\midrule
V-MAE$_{PT}$ - $L_2$ & 0.04 & 0.07 & 0.10 & -0.07\\
I3D - Entropy & -0.05 & -0.06 & -0.02 & 0.02\\
RAFT - Warp & \textbf{0.39} & \textbf{0.43} & \textbf{0.29} & 0.18 \\
MooG - PSNR & 0.28 & 0.32 & 0.26 & 0.09 \\
TRAJAN & \textbf{0.38} & 0.40 & 0.26 & \textbf{0.26} \\
\midrule
Inter-rater $\sigma$ & 0.09 & 0.07& 0.09& 0.12 \\
\bottomrule
\end{tabular}
\caption{\textbf{Validation Study.} Spearman's rank coefficients between human ratings and automated metrics for a subset of videos from EvalCrafter \citep{liu2024evalcrafter} (higher is better). Inter-rater $\sigma$ is the standard deviation of human responses (lower is better).}
\label{tab:human-consistency-realism-validation}
\end{table*}

\autoref{tab:human-speed-high} gives results for both EvalCrafter and VideoPhy when looking only at medium and high motion samples (those with human ratings of ``medium'' or ``high'' on the camera motion question). In these cases,  \model provides better correlations to human ratings, and humans are more consistent with each other (lower inter-rater $\sigma$s). This suggests, similar to the results on the external human datasets for EvalCrafter, that \model is capturing human ratings of consistency beyond just overall amount of motion.

\begin{figure*}

    \centering
    \begin{subfigure}[t]{0.9\textwidth}
    \includegraphics[width=\textwidth]{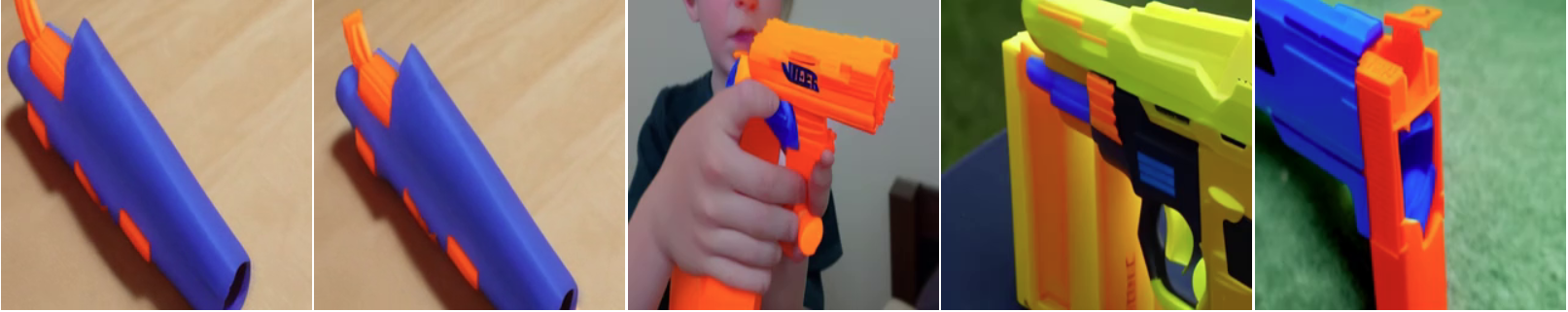}
    \caption{Example from the VideoPhy dataset sampled every 5 frames. This sample is scored in the upper half of the dataset for realism and appearance consistency by humans, and receives a physical consistency score of 1 from the original VideoPhy dataset. }
    \label{fig:videophy-choppy}
\end{subfigure}
        \centering
        \begin{subfigure}[t]{0.9\textwidth}
        \includegraphics[width=\textwidth]{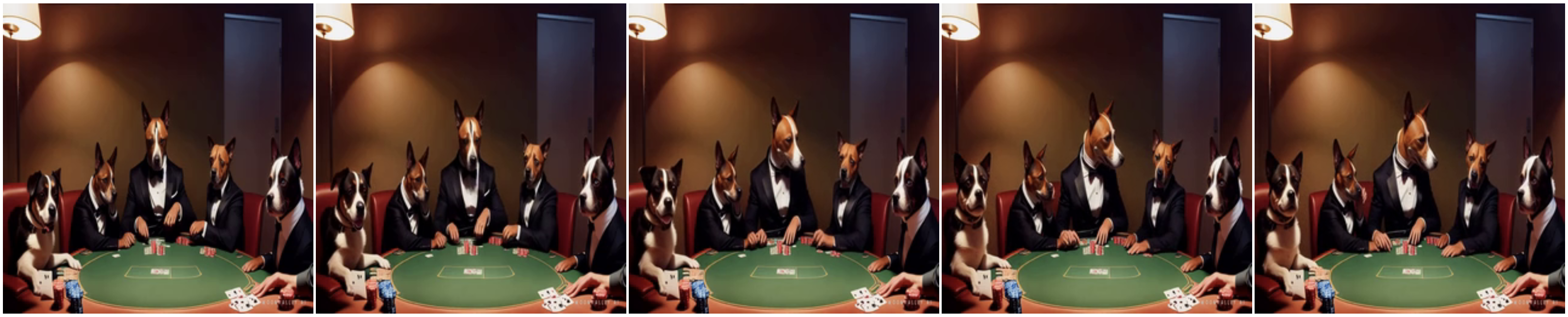}
        \caption{Humans rate badly on all consistency / realism metrics. All automated metrics score this highly. In this case, motion consistency and appearance consistency qualitatively appear good, but people seem unable to separate realism from consistency.}
    \end{subfigure}
    
    \begin{subfigure}[t]{0.9\textwidth}
        \centering
        \includegraphics[width=\textwidth]{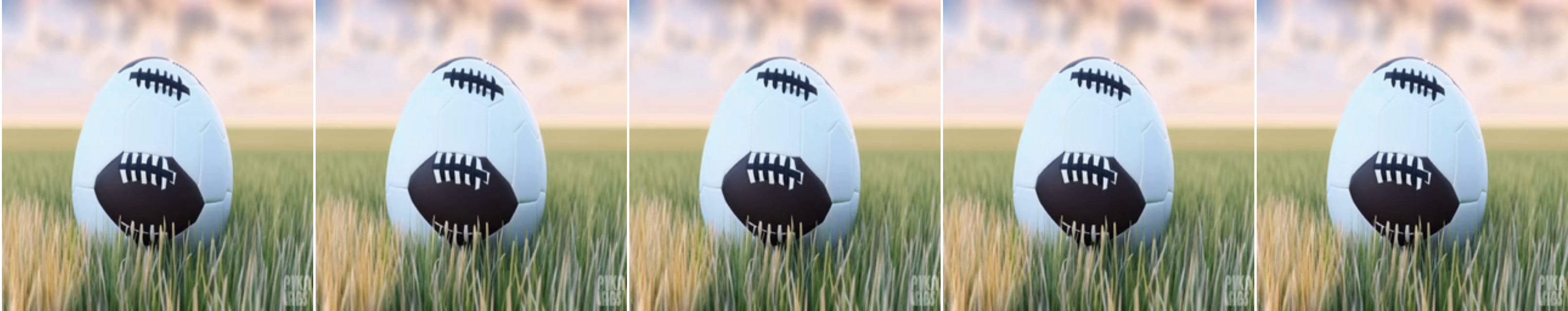}
        \caption{Humans rate badly on realism metrics. All automated metrics score this highly. In this case, motion consistency and appearance consistency qualitatively appear good, but people seem unable to separate the realism of the soccer / football from consistency.}
        \end{subfigure}
        \captionsetup{justification=centering}
        \caption{Examples from VideoPhy and EvalCrafter with human ratings.}
        \label{fig:videophy-samples}
    \end{figure*}

We also provide further qualitative examples of how people rate videos from the VideoPhy \citep{bansal2024videophy} and EvalCrafter \citep{liu2024evalcrafter} datasets in our human study in \autoref{fig:videophy-samples}. In \autoref{fig:videophy-samples}a, people rate the video as realistic, despite the fact that the video is temporally inconsistent. In \autoref{fig:videophy-samples}b, people rate the video as unrealistic and inconsistent, even though the motion of the dogs playing poker looks reasonable. In \autoref{fig:videophy-samples}c, people also rate the video as both unrealistic and inconsistent in appearance, even though the appearance does not change. People therefore seem strongly affected by the semantics of the objects even when prompted to ignore this information.

\paragraph{Validation Study.} We validate the UI for our human study by conducting a second human study on the same subset of videos obtained from the EvalCrafter dataset.
For this study, we replaced the slider scales with preset answer options corresponding to a 5-point Likert scale. Here option 1 implies a video that is very unrealistic (for questions about realism) or inconsistent (for questions about consistency) along this dimension, and option 5 implies a video that is very realistic (for questions about realism) or very consistent (for questions about consistency).
For this study we used 5 raters per question.

Results are reported in \autoref{tab:human-consistency-realism-validation}, where we post-processes in the same way as was done for \autoref{tab:human-consistency-realism}.
To make the results more comparable, we mapped the 1-5 Likert scale onto the slider scale by interpreting each scale as increments of 20.
It can be seen how TRAJAN and RAFT best correlate human judgement similar to in \autoref{tab:human-consistency-realism} (left) for consistency-related categories and realism, while TRAJAN performs clearly better for object interactions.

\section{Experiment Details}
\label{app:experimental-details}

\subsection{Datasets}

\subsubsection{UCF101}

For our experiments with synthethic distortions, we make use of the UCF101 dataset~\citep{soomro2012ucf101}, which consists of 13,320 videos recorded in the wild that show humans performing different types of actions.
We make use of the version available at \url{https://www.tensorflow.org/datasets/catalog/ucf101} having $256\times256$ resolution.  
Similar to prior work~\citep{ge2024content}, we obtain the `ground-truth' reference distribution by combining the first 32 frames of videos in the train and test split.

To obtain distorted videos, we consider the \emph{elastic transformation} from \citet{ge2024content}.
Details were obtained via personal correspondence with the authors, which we reproduce in the following.

\paragraph{Elastic Transformation~\citep{ge2024content}}
The reference implementation for this transformation is \url{https://github.com/hendrycks/robustness/blob/master/ImageNet-C/imagenet_c/imagenet_c/corruptions.py}, which works by first performing an affine- and then an elastic transformation.
The reference implementation suggests five levels of degradation~\citep{hendrycks2018benchmarking}.
The first two levels (referenced as 1.1 and 1.2 in \autoref{fig:corruptions}) primarily trade-off the strength of the affine transformation with that of the elastic transform, such that 1.2 creates more global distortions to the shape of objects.
Levels 2.1, 2.2, and 2.3 fix the affine transformation at small intensity, and only increase the strength of the elastic part.
This results in much more local distortions associated with high-frequency noise. 
\citet{ge2024content} adopt these sample levels, except that they adjust the resolution parameter from $244$ to $128$.
Here we adopt the same distortion levels as in \citet{ge2024content}. 

To compute the temporal sensitivity of a metric, \citet{ge2024content} proposes to evaluate them on both \emph{spatially} distorted videos and on \emph{spatiotemporally} distorted videos, and report the ratio of the two.
In practice, the elastic transformation is applied at a per-frame level, where the distortion levels parametrize a distribution of corruptions that are sampled from.
When using the same seed between frames in a video, the same corruption is applied to each frame, creating only a spatial effect.
To obtain a spatiotemporal effect, a different corruption is sampled for each frame in the video.

\begin{figure*}
    \captionsetup[subfigure]{justification=centering}
    \centering
    \begin{subfigure}[t]{0.9\textwidth}
        \centering
        \includegraphics[height=1.2in]{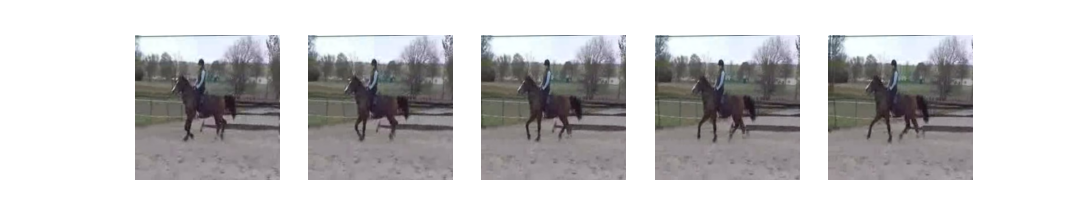}
        \caption{Spatial-only corruptions - level 1.1}
    \end{subfigure}
    ~
        \begin{subfigure}[t]{0.9\textwidth}
        \centering
        \includegraphics[height=1.2in]{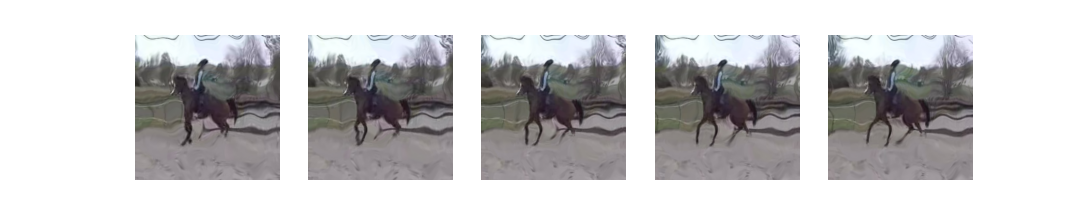}
        \caption{Spatial-only corruptions - level 1.2}
    \end{subfigure}
    ~
        \begin{subfigure}[t]{0.9\textwidth}
        \centering
        \includegraphics[height=1.2in]{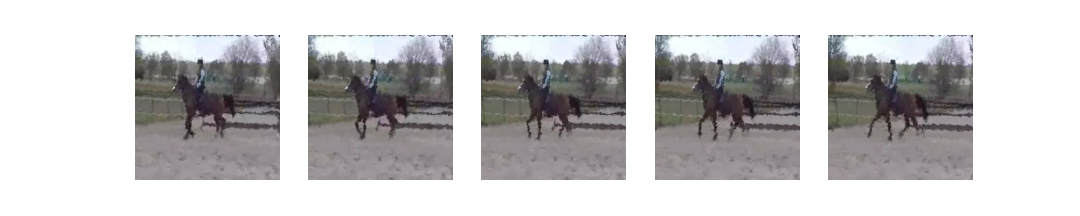}
        \caption{Spatial-only corruptions - level 2.1}
    \end{subfigure}
    ~
        \begin{subfigure}[t]{0.9\textwidth}
        \centering
        \includegraphics[height=1.2in]{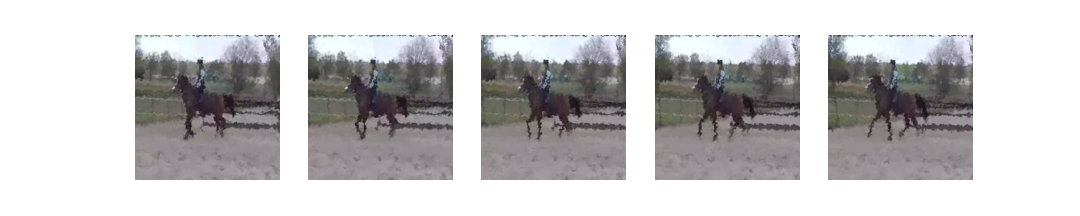}
        \caption{Spatial-only corruptions - level 2.2}
    \end{subfigure}
    ~
        \begin{subfigure}[t]{0.9\textwidth}
        \centering
        \includegraphics[height=1.2in]{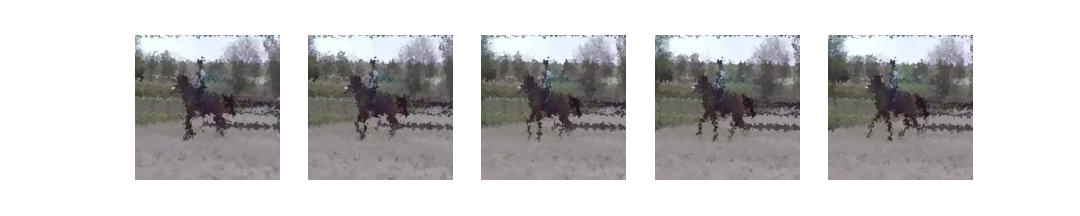}
        \caption{Spatial-only corruptions - level 2.3}
    \end{subfigure}
    \caption{A UCF-101 video with \emph{spatial} corruptions using the five levels of elastic transformation from \citet{ge2024content}.}
    \label{fig:ucf101-spatial}
    \end{figure*}

\begin{figure*}
    \captionsetup[subfigure]{justification=centering}
        \begin{subfigure}[t]{0.9\textwidth}
        \centering
        \includegraphics[height=1.2in]{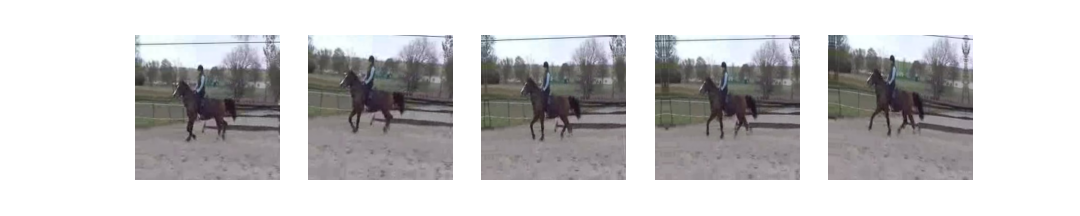}
        \caption{Spatiotemporal corruptions - level 1.1}
    \end{subfigure}
    ~
        \begin{subfigure}[t]{0.9\textwidth}
        \centering
        \includegraphics[height=1.2in]{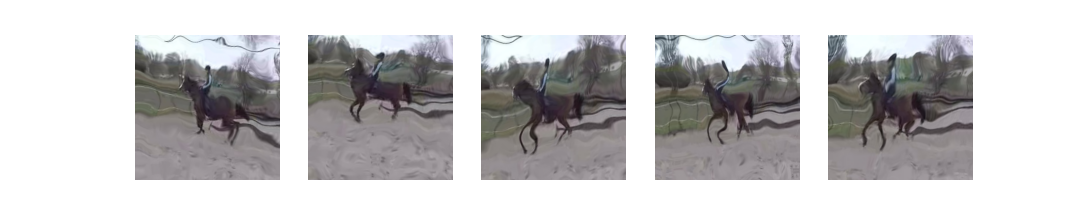}
        \caption{Spatiotemporal corruptions - level 1.2}
    \end{subfigure}
    ~
        \begin{subfigure}[t]{0.9\textwidth}
        \centering
        \includegraphics[height=1.2in]{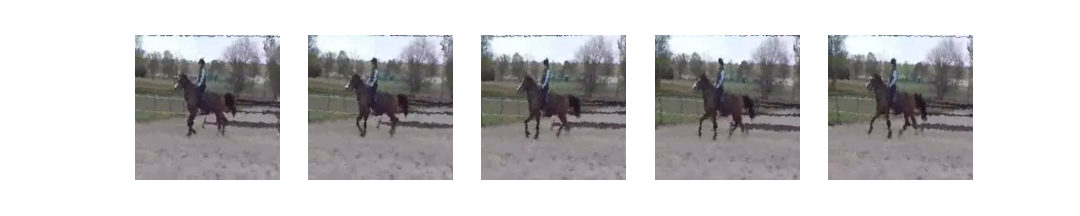}
        \caption{Spatiotemporal corruptions - level 2.1}
    \end{subfigure}
    ~
        \begin{subfigure}[t]{0.9\textwidth}
        \centering
        \includegraphics[height=1.2in]{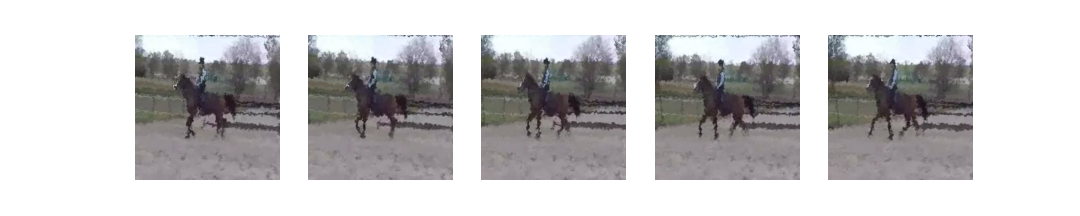}
        \caption{Spatiotemporal corruptions - level 2.2}
    \end{subfigure}
    ~
        \begin{subfigure}[t]{0.9\textwidth}
        \centering
        \includegraphics[height=1.2in]{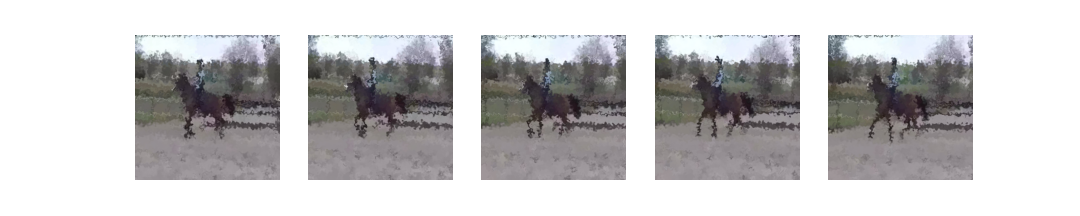}
        \caption{Spatiotemporal corruptions - level 2.3}
    \end{subfigure}
    \caption{A UCF-101 video with \emph{spatiotemporal} corruptions using the five levels of elastic transformation from \citet{ge2024content}.}
    \label{fig:ucf101-spatiotemporal}
\end{figure*}

In Figures \ref{fig:ucf101-spatial} \& \ref{fig:ucf101-spatiotemporal} we show an example of a UCF-101 video with spatial- or spatio-temporal corruptions applied using the five levels of elastic transformation from \citet{ge2024content}.
Note that in the spatial setting, the same corruption is applied to each frame, while in the spatiotemporal setting the corruption parameters are resampled for each frame.


\subsubsection{VideoPhy}

We make use of the ``solid-solid'' split of VideoPhy~\citep{bansal2024videophy} available for download at \url{https://huggingface.co/datasets/videophysics/videophy_test_public}.
These include generated videos from several competitive models, including CogVideoX (2b and 5b)~\citep{yang2024cogvideox}, Gen-2~\citep{esser2023structure}, LaVIE~\citep{wang2023lavie}, OpenSora~\citep{zheng1open}, Pika~\citep{pika}, SVD~\citep{blattmann2023stable}, VideoCrafter2~\citep{chen2024videocrafter2}, and ZeroScope~\citep{zeroscope}.
Note that at the time of preparing this submission, we were unable to download videos contained in the dataset belonging to Luma Dream Machine, which were made available at a later date.
We pre-processed all videos to $256\times256$ resolution.

\subsubsection{EvalCrafter}
The EvalCrafter dataset~\citep{liu2024evalcrafter} contains generated videos for a diverse set of prompts and is available for download at \url{https://huggingface.co/datasets/RaphaelLiu/EvalCrafter_T2V_Dataset}.
To obtain a comparable size to VideoPhy, while preserving diversity as much as possible, we randomly select 104 generated videos from 11 of the text-to-video models (the original 5 from the EvalCrafter human evaluation dataset, and 6 additional models). 
These models are:
\begin{enumerate}
    \item Gen2 (December) \citep{esser2023structure}
     \item MoonValley \citep{moonvalley}
      \item PikaLab (December) \citep{pika}
      \item Show-1 \citep{zhang2024show}
      \item VideoCrafter 1.0 \citep{chen2023videocrafter1}
      \item Hotshot-XL \citep{hotshot}
      \item PikaLab \citep{pika}
      \item Gen2  \citep{esser2023structure}
      \item Floor33 Pictures \citep{floor33}
      \item ZeroScope \citep{zeroscope}
      \item ModelScope \citep{wang2023modelscope}
\end{enumerate}
We obtain 1144 videos in this way, which we preprocess to $256\times256$ resolution.

\subsubsection{WALT Generations}
We train a WALT~\citep{gupta2025photorealistic} diffusion model (214M params) for frame-conditional video generation on the Kinetics-600 dataset~\citep{carreira2018short}. The model architecture and hyperparameters are consistent with those used in the original paper. Training is conducted for 495,000 iterations with a batch size of 256, using videos at a resolution of $128 \times 128$ with 17 temporal frames. Following WALT, the model is designed to predict three future latent frames conditioned on the initial two latent frames. To evaluate performance, we compute metrics at key checkpoints, using model weights saved at steps $1$, $100,000$, $200,000$, $300,000$, $400,000$, and $495,000$.
We upsample generated videos to $256\times256$ resolution and make use of the first 16 frames.
We show the distribution level metrics and single video metrics evaluated on samples generated at these intermediate steps in~\Cref{fig:walt_training}.

\subsection{Models}
\label{supp:models}
\subsubsection{\model}
\label{supp:trajan}

We train \model on a video dataset of publicly accessible videos following~\citet{doersch2024bootstap}, aiming for high-quality and realistic motion. We use videos tagged as lifestyle and one-shot videos, but omit those videos from categories with low
visual complexity or unrealistic motions, e.g. tutorial videos, lyrics videos,
and animations.  We select only 60fps videos with
over 200 views, without cuts or overlays. However, a key difference from prior work~\citep{doersch2024bootstap} is that we use 150-frame clips, whereas prior work focused on 24-frame clips for bootstrapping point tracking.  We sample 15 million such videos, and for each video, we choose 4096 points uniformly at random across space and time for every clip and track them using the public BootsTAPIR~\citep{doersch2024bootstap} model.

There are three key challenges for representing a set of point tracks as a dense vector where distances are meaningful.  1) point tracks are \textit{orderless}, in the sense that they cannot be arranged into a sequence of tokens: unlike image pixels, tracks do not occur on a grid, and there may be no single video frame where all tracks are visible.  Therefore, we wish to have a permutation-invariant representation.  2) occlusions should be treated as missing data, rather than as a fundamental part of the motion, since two motions may be very similar in the real world even if one is partially occluded.  Finally, 3) although we are representing a dense signal (every point in the image corresponds to a single track), for computational reasons we only receive a finite number of samples from the underlying function.  The exact points that are chosen for tracking are not important for representing motion, so we aim for our representation to be invariant to the specific chosen points.

Let $S=\{s_{t,j}\}$ be point trajectories and occlusions, where $s_{t,j}\in \mathbb{R}^{3} = (x_{t,j},y_{t,j},o_{t,j})$ corresponds to $x$ and $y$ positions and occlusion flag $o$ at time $t$ for the $j$th track.
As mentioned, we embed all $(x_{t,j},y_{t,j},t)$, add a readout token, and perform self-attention using $(1-o_{t,j})$ as an attention mask, which helps us achieve point 1) above, as the transformer sees these points as `masked' rather than some dummy value.  After self attention, we discard all tokens except the ``readout'' token, which now provides a fixed-length $C$-channel representation of each track.  We then apply a Perceiver~\citep{jaegle2021perceiver} to encode all track tokens; we apply no additional position encoding, meaning that the entire representation is permutation invariant, following principle 1) above.  Finally, we project the latent tokens to a lower dimension resulting in a fixed-size $128 \times 64$ dimensional representation $\phi_S$ of the tracks.

As mentioned, we achieve point 3) above by decoding points that are \textit{not} included in the autoencoder input.  The autoencoder knows which track to decode because we give a query point $x_q,y_q,t_q$ on any frame, and we output the track that goes through this point.  Note that this means that the autoencoder can actually output truly \textit{dense} motion information even though it receives only a finite sample as input.  As stated, we up-project the tokens in $\phi_S$ to a higher dimension with an operator $U$, apply a transformer with a readout token, and apply a linear mapping to occlusion logit $o^q_t$ and $x^q_t,y^q_t$.  We train them with softmax cross entropy and Huber loss, respectively, with a weight of $5000$ on the Huber loss and $1e-8$ on the cross entropy; the disparity between loss weights is due to the fact that we mostly want the representation to be mostly invariant to occlusion and focus on motion. In initial experiments, we found that setting these weights equally led to worse performance in correlating with human judgements of realism in generated videos. We find that a naive linear up-projection operator $U$ tends to result in poor temporal localization of the query point, as the model struggles to cross-attend to the correct latent tokens.  We find that we can improve performance by using an upsampling operator that first linearly up-projects, and then extracts a window of each token, concatenating along the channel axis.  That is, the up-projected representation for the $l$th token is $U(\phi_S^l)=concat(\left[f(\phi_S^l),f(\phi_S^l)[\rho t:\rho t + 128\right])$, where $f$ is a linear projection, $[\cdot]$ represents indexing, and $\rho$ is a stride.  This can be seen as specializing the motion tokens for time $t$, so the model can more easily identify what motion information is relevant to this query.  We train with Adam~\citep{adam} with a warmup cosine learning rate schedule with 1000 warmup steps and a peak learning rate of 2e-4 for 1M steps with a batch size of 64.  

Note that not all of the videos we would like to evaluate have 150 frames of motion.  Therefore, we train the model to also encode shorter clips.  Given motion from a 150-frame clip, for half of the examples, we (uniformly) randomly sample an `end' frame, and mark all points after a certain length as `occluded'.  For these examples, we only apply the loss to the frames before the `end' frame.

We evaluate reconstruction accuracy with Average Jaccard, following TAP-Vid~\citep{doersch2022tap}.  Given a threshold $\delta$ ``true positives'' ($TP$) are predictions which are within $\delta$ of the ground truth.  ``False positives'' ($FP$) are predictions that are farther than $\delta$ from the ground truth (or the ground truth is occluded), and ``false negatives'' ($FN$) are ground truth points where the prediction is farther than $\delta$ (or occluded).  $\mbox{Jaccard}_{\delta}$ is $TP/(TP+FP+FN)$, and Average Jaccard averages $\mbox{Jaccard}_{\delta}$ over several pixel thresholds.  We use the same thresholds proposed in TAP-Vid~\citep{doersch2022tap}: namely, we resize all trajectories as if they had come from a $256 \times 256$ video, and use thresholds of 1, 2, 4, 8, and 16 pixels.  We find that this model is quite accurate for real-world training data, with average points within threshold of 85.3 when evaluated on held-out data from the same distribution, and an average jaccard of 55.8, roughly meeting the performance of the underlying BootsTAPIR tracker.

\paragraph{Architecture details}

For our transformer implementation, we use the standard design from \citet{transformer} with the pre-layer norm configuration from \citet{xiong2020layer}, and some of the additional improvements introduced in \citet{dehghani2023scaling}. Specifically, we use the RMS norm applied to the keys and queries before computing attention weights, and execute self- and cross-attention paths (not the MLP path) in parallel. Another Layer Normalization layer \citep{layernorm} is applied to the output. Transformer hyperparameters are given in \autoref{tab:transformerhypers} with names given by their corresponding descriptions in the main text.

Additional hyperparameters for the sinusoidal positional embeddings, and the dimensionalities of projection operators, are given in \autoref{tab:projectionhypers}.

\begin{table*}[h]
    \small
    \centering
    \begin{tabular}{l|c}
    \toprule
      Sinusoidal embedding (number of frequencies) & 32 \\
      Track token projection dimensionality ($C$) & 256 \\
      Compression dimensionality & 64 \\
      Up-projection dimensionality & 1024 - 128 \\
      Query point encoder dimensionality & 1024 \\
      \bottomrule
    \end{tabular}
    \captionsetup{justification=centering}
    \caption{Positional encoding and projection operator hyperparameters for \model.}
    \label{tab:projectionhypers}
\end{table*}

\begin{table*}[h]
    \centering
    \small
    \begin{tabular}{lccccc}
    \toprule
    Transformer name & Attention type & QKV size & Layers & Heads & MLP size \\
    \toprule
    Input track transformer & SA & 64$\times$8 & 2 & 8 & 1024 \\
    Perceiver-style tracks to latents & CA & 64$\times$8 & 3 & 8 & 2048 \\
    Up-projection latent transformer in decoder & CA & 64$\times$8 & 3 & 8 & 2048 \\
    Track readout transformer & CA & 64$\times$8 & 4 & 8 & 1024 \\
    \bottomrule
    \end{tabular}
    \caption{Transformer architecture hyperparameters for \model. SA = self-attention, CA = cross-attention.}
    \label{tab:transformerhypers}
\end{table*}

\begin{figure}
    \centering
    \includegraphics[scale=0.4]{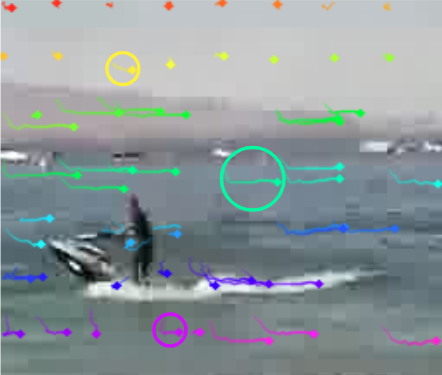}
    \caption{Examples of point trajectory radii. Circles are shown around maximum amount of motion. Radii are calculated from these circles.}
    \label{fig:trackradii}
\end{figure}

\paragraph{Track motion radii calculation}
From the point tracks output by BootsTAPIR \citep{doersch2024bootstap}, we can calculate two metrics of their overall motion: (1) their track lengths ($\sum_t \sqrt{(x_{t+1}-x_t)^2 + (y_{t+1}-y_t)^2}$), masking out non-visible tracks, and (2) their track radii (\autoref{fig:trackradii}), calculated as the radius of the smallest enclosing circle from the start to the end of a visible track. The track lengths will cover jittery movements or objects changing direction, while the motion radii will capture the maximum distance, in a consistent direction, for each point.

\subsubsection{Motion Histograms}
Our implementation of motion histograms~\citep{liu2024frechet} is based on \url{https://github.com/DSL-Lab/FVMD-frechet-video-motion-distance}.
They propose to evaluate motion consistency in generated videos by estimating velocity and acceleration from point tracks.
Inspired by HOG features~\citep{dalal2005histograms}, they partition the resulting volumes into $4\times5\times5$-sized tubelets and accumulate the magnitude of the values at each angle (using 8 bins) within a tubelet.
Motion features are obtained by concatenating the resulting 1D histograms obtained for each tubelet using both velocity and acceleration. 
We apply this approach to 16-frame chunks of $64\times64$ densely sampled point tracks obtained from BootsTAPIR~\citep{doersch2024bootstap} to yield a 9216-dimensional vector describing the motion within the corresponding 16-frame video.

\subsubsection{RAFT}

To estimate optical flow we use the improved version~\citep{sun2022disentangling} of RAFT~\citep{teed2020raft} used as a baseline in \citet{saxena2024surprising}.
Since we are only concerned with obtaining the best possible flow model, we use a RAFT model which was trained on a large mixture of standard optical flow datasets.
These include Sintel~\citep{Butler:ECCV:2012}, KITTI~\citep{geiger2013vision}, Kubric~\citep{greff2022kubric}, TartanAir~\citep{wang2020tartanair}, FlyingThings~\citep{mayer2016large}, and AutoFlow~\citep{sun2021autoflow}.

\subsubsection{MooG}

MooG~\citep{vanSteenkiste2024moving} is a recurrent model trained for next-frame prediction. It operates by maintaining and updating an internal state constituting of off-the-grid latent tokens, which can be decoded to predict the next frame. These latents are first randomly intialized, and then updated on each iteration by a transformer model which cross attends to the image features of the corresponding frame, followed by a set of self-attention layers. The decoder converts the latent state back to pixels by querying the latents through cross-attention with fixed grid-based features~\citep{sajjadi2022scene,jaegle2022perceiver}.

The original MooG implementation distinguishes between a ``predicted'' state and a ``corrected'' state, where the former only depends on the previous state, while the latter additionally integrates the current observation.
To avoid the model only reconstructing the current observation, the loss is placed on the predicted state in this case.
Here we slightly simplify the implementation by using placing the loss on the corrected state, but using the \emph{next} observation when computing the loss.
We train MooG for 600K steps on a mixture of datasets, including Kinetics-700~\citep{carreira2018short}, SSv2~\citep{goyal2017something}, ScanNet~\citep{dai2017scannet}, Ego4D~\citep{grauman2022ego4d}, and Walking Tours~\citep{venkataramanan2024imagenet}.

\subsection{Human Study}
\label{app:experimental-details-human-study}

To evaluate whether the proposed metrics are useful, we compare them to human judgements.
Prior works mainly source human labels at a coarse-grained level using pair-wise comparisons between videos.
For example, \citet{unterthiner2018towards,luo2024beyond} ask humans to compare videos from two different sources (either different models, or with different levels of noise applied to them) to determine which of the two looked better, or alternatively report that their quality was indistinguishable.
\citet{liu2024frechet} take a similar approach using three pairwise comparisons: if a human expresses the same preference for at least two of the pairs then they are determined to prefer that model.

An alternative is to ask humans to evaluate individual videos as in \citet{kim2024stream}. 
There they use a 6-point scale to evaluate realism of videos, and a 3-point scale for motion or ``temporal naturalness''.
Similarly, \citet{bansal2024videophy} asks humans to indicate with yes/no  whether videos ``follow Physics Laws or Physical Commonsense''.
\citet{liu2024evalcrafter} asks humans to score individual videos based on their ``motion quality'' and ``temporal consistency'' using a 5-point scale. 

\begin{figure}
    \centering
    \includegraphics[width=\linewidth]{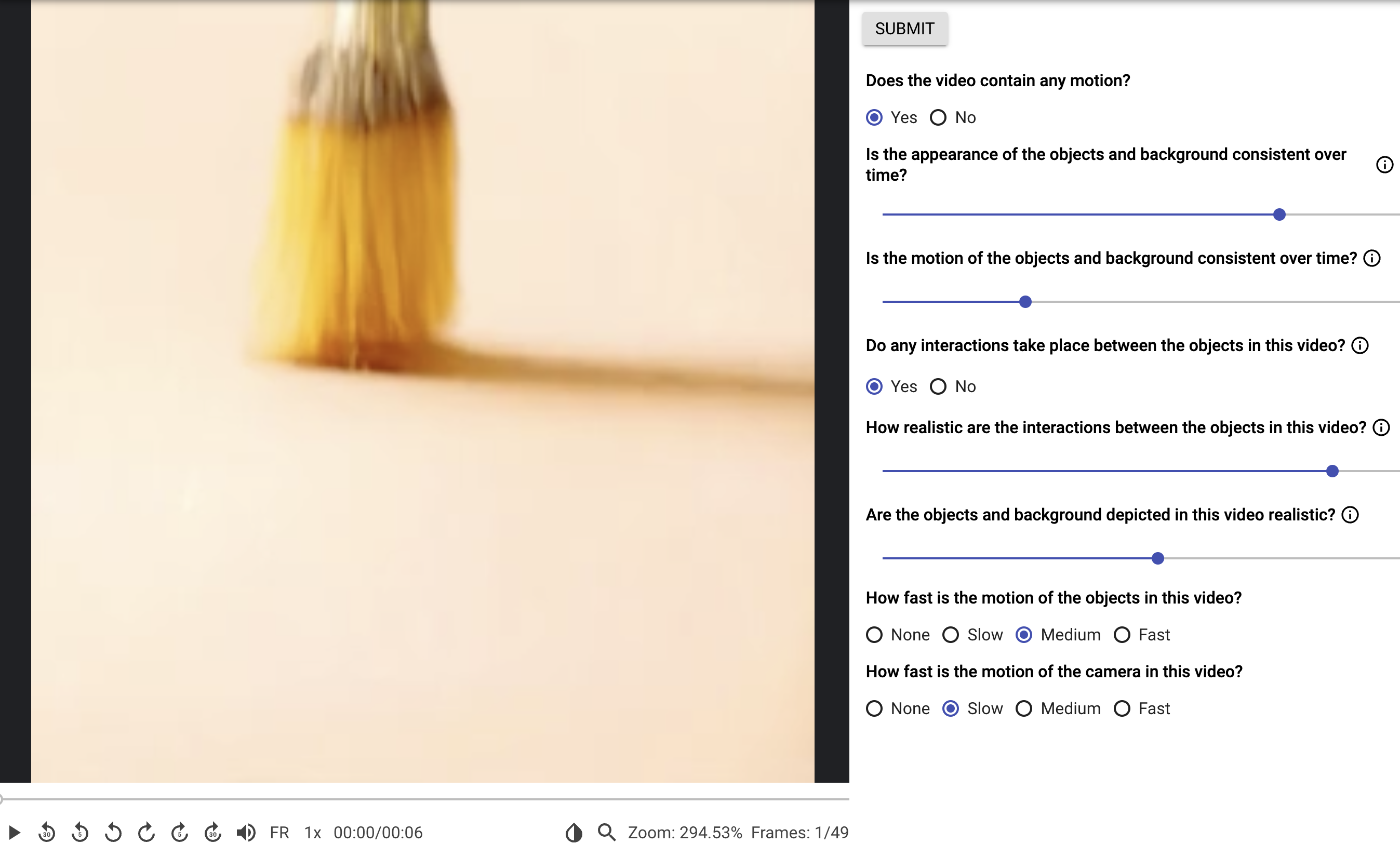}
    \caption{\textbf{A screenshot of the rater UI}. The video plays automatically on repeat as soon as the UI is opened. Videos were rendered at 8fps for ease of rating.}
    \label{fig:rater}
\end{figure}

A possible concern when sourcing human labels in this way is that the questions might be too open-ended, which limits their usefulness for evaluating motion in generated videos and for developing corresponding metrics.
More generally, it isn't clear what motion-related dimensions are measured when asking about ``physical commonsense'' or when having humans express a preference for one generated video over another.
To improve upon this, we propose to source fine-grained human annotations for generated videos using the questions detailed in \autoref{fig:rater}. 
To avoid conflating perfect consistency in videos that have motion with videos that contain no motion, we first ask humans about whether a video contains any motion at all, and only follow-up with the next two consistency related questions if they answer ``Yes''.
Similarly, we first ask humans about whether interactions take place in a video.
If they answer ``Yes'', then the slider for rating how realistic these interactions are appears.
The other questions are available at all times and not conditioned on any of the previous answers.

We made use of a rater pool of 10 raters, and used 3 raters per question (randomly assigned).
Before conducting the full human study, we asked raters to rate 10 questions, and gave feedback on their responses.
For example, after an initial pilot, we noticed that many of the ratings were either 0 or 100 (both extreme endpoints of the slider), with few responses in between, and encouraged raters to make use of the full range of the slider to express degrees of agreement.
We also noticed that camera motion was initially incorrectly evaluated, which we clarified.
Feedback was also provided once we had obtained the full results for VideoPhy, before starting EvalCrafter.
In particular, we noted that many videos with inconsistent motion (such as the objects or camera jittering / jumping around) received high scores for motion consistency.
This is not desirable as for this data it can be assumed that the objects / camera are expected to move in a smooth and continuous manner, which we clarified.
We encountered several videos with plausible looking individual frames, but that are stitched together in unnatural ways (no continuity between the frames in terms of content, see \autoref{fig:videophy-choppy}), which received high realism scores.
In this case we provided feedback that this question concerns the entire video as a whole and not the sum of individual frames.

%% file: main.bbl
\begin{thebibliography}{89}
\providecommand{\natexlab}[1]{#1}
\providecommand{\url}[1]{\texttt{#1}}
\expandafter\ifx\csname urlstyle\endcsname\relax
  \providecommand{\doi}[1]{doi: #1}\else
  \providecommand{\doi}{doi: \begingroup \urlstyle{rm}\Url}\fi

\bibitem[Ba et~al.(2016)Ba, Kiros, and Hinton]{layernorm}
J.~L. Ba, J.~R. Kiros, and G.~E. Hinton.
\newblock Layer normalization.
\newblock \emph{{NeurIPS Deep Learning Symposium}}, 2016.

\bibitem[Bansal et~al.(2024{\natexlab{a}})Bansal, Bitton, Szpektor, Chang, and
  Grover]{bansal2024videocon}
H.~Bansal, Y.~Bitton, I.~Szpektor, K.-W. Chang, and A.~Grover.
\newblock Videocon: Robust video-language alignment via contrast captions.
\newblock In \emph{IEEE Conf. Comput. Vis. Pattern Recog.}, pages 13927--13937,
  2024{\natexlab{a}}.

\bibitem[Bansal et~al.(2024{\natexlab{b}})Bansal, Lin, Xie, Zong, Yarom,
  Bitton, Jiang, Sun, Chang, and Grover]{bansal2024videophy}
H.~Bansal, Z.~Lin, T.~Xie, Z.~Zong, M.~Yarom, Y.~Bitton, C.~Jiang, Y.~Sun,
  K.-W. Chang, and A.~Grover.
\newblock Videophy: Evaluating physical commonsense for video generation.
\newblock \emph{arXiv preprint arXiv:2406.03520}, 2024{\natexlab{b}}.

\bibitem[Bi{\'n}kowski et~al.(2018)Bi{\'n}kowski, Sutherland, Arbel, and
  Gretton]{binkowski2018demystifying}
M.~Bi{\'n}kowski, D.~J. Sutherland, M.~Arbel, and A.~Gretton.
\newblock Demystifying mmd gans.
\newblock In \emph{Int. Conf. Learn. Represent.}, 2018.

\bibitem[Blattmann et~al.(2023{\natexlab{a}})Blattmann, Dockhorn, Kulal,
  Mendelevitch, Kilian, Lorenz, Levi, English, Voleti, Letts,
  et~al.]{blattmann2023stable}
A.~Blattmann, T.~Dockhorn, S.~Kulal, D.~Mendelevitch, M.~Kilian, D.~Lorenz,
  Y.~Levi, Z.~English, V.~Voleti, A.~Letts, et~al.
\newblock Stable video diffusion: Scaling latent video diffusion models to
  large datasets.
\newblock \emph{arXiv preprint arXiv:2311.15127}, 2023{\natexlab{a}}.

\bibitem[Blattmann et~al.(2023{\natexlab{b}})Blattmann, Rombach, Ling,
  Dockhorn, Kim, Fidler, and Kreis]{blattmann2023align}
A.~Blattmann, R.~Rombach, H.~Ling, T.~Dockhorn, S.~W. Kim, S.~Fidler, and
  K.~Kreis.
\newblock Align your latents: High-resolution video synthesis with latent
  diffusion models.
\newblock In \emph{IEEE Conf. Comput. Vis. Pattern Recog.}, pages 22563--22575,
  2023{\natexlab{b}}.

\bibitem[Brooks et~al.(2024)Brooks, Peebles, Holmes, DePue, Guo, Jing, Schnurr,
  Taylor, Luhman, Luhman, Ng, Wang, and Ramesh]{videoworldsimulators2024}
T.~Brooks, B.~Peebles, C.~Holmes, W.~DePue, Y.~Guo, L.~Jing, D.~Schnurr,
  J.~Taylor, T.~Luhman, E.~Luhman, C.~Ng, R.~Wang, and A.~Ramesh.
\newblock Video generation models as world simulators.
\newblock 2024.
\newblock URL
  \url{https://openai.com/research/video-generation-models-as-world-simulators}.

\bibitem[Bugliarello et~al.(2024)Bugliarello, Moraldo, Villegas, Babaeizadeh,
  Saffar, Zhang, Erhan, Ferrari, Kindermans, and
  Voigtlaender]{bugliarello2024storybench}
E.~Bugliarello, H.~H. Moraldo, R.~Villegas, M.~Babaeizadeh, M.~T. Saffar,
  H.~Zhang, D.~Erhan, V.~Ferrari, P.-J. Kindermans, and P.~Voigtlaender.
\newblock Storybench: a multifaceted benchmark for continuous story
  visualization.
\newblock \emph{Adv. Neural Inform. Process. Syst.}, 36, 2024.

\bibitem[Butler et~al.(2012)Butler, Wulff, Stanley, and
  Black]{Butler:ECCV:2012}
D.~J. Butler, J.~Wulff, G.~B. Stanley, and M.~J. Black.
\newblock A naturalistic open source movie for optical flow evaluation.
\newblock In {A. Fitzgibbon et al. (Eds.)}, editor, \emph{ECCV}, Part IV, LNCS
  7577, pages 611--625. Springer-Verlag, Oct. 2012.

\bibitem[Carreira and Zisserman(2017)]{carreira2017quo}
J.~Carreira and A.~Zisserman.
\newblock Quo vadis, action recognition? a new model and the kinetics dataset.
\newblock In \emph{IEEE Conf. Comput. Vis. Pattern Recog.}, pages 6299--6308,
  2017.

\bibitem[Carreira et~al.(2018)Carreira, Noland, Banki-Horvath, Hillier, and
  Zisserman]{carreira2018short}
J.~Carreira, E.~Noland, A.~Banki-Horvath, C.~Hillier, and A.~Zisserman.
\newblock A short note about kinetics-600.
\newblock \emph{arXiv preprint arXiv:1808.01340}, 2018.

\bibitem[Chen et~al.(2023)Chen, Xia, He, Zhang, Cun, Yang, Xing, Liu, Chen,
  Wang, et~al.]{chen2023videocrafter1}
H.~Chen, M.~Xia, Y.~He, Y.~Zhang, X.~Cun, S.~Yang, J.~Xing, Y.~Liu, Q.~Chen,
  X.~Wang, et~al.
\newblock Videocrafter1: Open diffusion models for high-quality video
  generation.
\newblock \emph{arXiv preprint arXiv:2310.19512}, 2023.

\bibitem[Chen et~al.(2024)Chen, Zhang, Cun, Xia, Wang, Weng, and
  Shan]{chen2024videocrafter2}
H.~Chen, Y.~Zhang, X.~Cun, M.~Xia, X.~Wang, C.~Weng, and Y.~Shan.
\newblock Videocrafter2: Overcoming data limitations for high-quality video
  diffusion models.
\newblock In \emph{IEEE Conf. Comput. Vis. Pattern Recog.}, pages 7310--7320,
  2024.

\bibitem[Dai et~al.(2017)Dai, Chang, Savva, Halber, Funkhouser, and
  Nie{\ss}ner]{dai2017scannet}
A.~Dai, A.~X. Chang, M.~Savva, M.~Halber, T.~Funkhouser, and M.~Nie{\ss}ner.
\newblock Scannet: Richly-annotated 3d reconstructions of indoor scenes.
\newblock In \emph{IEEE Conf. Comput. Vis. Pattern Recog.}, pages 5828--5839,
  2017.

\bibitem[Dalal and Triggs(2005)]{dalal2005histograms}
N.~Dalal and B.~Triggs.
\newblock Histograms of oriented gradients for human detection.
\newblock In \emph{Proc. IEEE Comput. Soc. Conf. Comput. Vis. Pattern Recog.},
  volume~1, pages 886--893. Ieee, 2005.

\bibitem[Dehghani et~al.(2023)Dehghani, Djolonga, Mustafa, Padlewski, Heek,
  Gilmer, Steiner, Caron, Geirhos, Alabdulmohsin, et~al.]{dehghani2023scaling}
M.~Dehghani, J.~Djolonga, B.~Mustafa, P.~Padlewski, J.~Heek, J.~Gilmer, A.~P.
  Steiner, M.~Caron, R.~Geirhos, I.~Alabdulmohsin, et~al.
\newblock Scaling vision transformers to 22 billion parameters.
\newblock In \emph{Int. Conf. Mach. Learn.}, pages 7480--7512. PMLR, 2023.

\bibitem[Doersch et~al.(2022)Doersch, Gupta, Markeeva, Recasens, Smaira, Aytar,
  Carreira, Zisserman, and Yang]{doersch2022tap}
C.~Doersch, A.~Gupta, L.~Markeeva, A.~Recasens, L.~Smaira, Y.~Aytar,
  J.~Carreira, A.~Zisserman, and Y.~Yang.
\newblock {TAP}-vid: A benchmark for tracking any point in a video.
\newblock \emph{Adv. Neural Inform. Process. Syst.}, 35:\penalty0 13610--13626,
  2022.

\bibitem[Doersch et~al.(2024)Doersch, Luc, Yang, Gokay, Koppula, Gupta,
  Heyward, Rocco, Goroshin, Carreira, et~al.]{doersch2024bootstap}
C.~Doersch, P.~Luc, Y.~Yang, D.~Gokay, S.~Koppula, A.~Gupta, J.~Heyward,
  I.~Rocco, R.~Goroshin, J.~Carreira, et~al.
\newblock {BootsTAP: Bootstrapped training for tracking-any-point}.
\newblock In \emph{Asian Conf. Comput. Vis.}, 2024.

\bibitem[Esser et~al.(2023)Esser, Chiu, Atighehchian, Granskog, and
  Germanidis]{esser2023structure}
P.~Esser, J.~Chiu, P.~Atighehchian, J.~Granskog, and A.~Germanidis.
\newblock Structure and content-guided video synthesis with diffusion models.
\newblock In \emph{Int. Conf. Comput. Vis.}, pages 7346--7356, 2023.

\bibitem[Ge et~al.(2024)Ge, Mahapatra, Parmar, Zhu, and Huang]{ge2024content}
S.~Ge, A.~Mahapatra, G.~Parmar, J.-Y. Zhu, and J.-B. Huang.
\newblock On the content bias in fr{\'e}chet video distance.
\newblock In \emph{IEEE Conf. Comput. Vis. Pattern Recog.}, pages 7277--7288,
  2024.

\bibitem[Geiger et~al.(2013)Geiger, Lenz, Stiller, and
  Urtasun]{geiger2013vision}
A.~Geiger, P.~Lenz, C.~Stiller, and R.~Urtasun.
\newblock Vision meets robotics: The kitti dataset.
\newblock \emph{Int. J. Robotics Res.}, 32\penalty0 (11):\penalty0 1231--1237,
  2013.

\bibitem[Goyal et~al.(2017)Goyal, Ebrahimi~Kahou, Michalski, Materzynska,
  Westphal, Kim, Haenel, Fruend, Yianilos, Mueller-Freitag,
  et~al.]{goyal2017something}
R.~Goyal, S.~Ebrahimi~Kahou, V.~Michalski, J.~Materzynska, S.~Westphal, H.~Kim,
  V.~Haenel, I.~Fruend, P.~Yianilos, M.~Mueller-Freitag, et~al.
\newblock The" something something" video database for learning and evaluating
  visual common sense.
\newblock In \emph{Int. Conf. Comput. Vis.}, pages 5842--5850, 2017.

\bibitem[Grauman et~al.(2022)Grauman, Westbury, Byrne, Chavis, Furnari,
  Girdhar, Hamburger, Jiang, Liu, Liu, et~al.]{grauman2022ego4d}
K.~Grauman, A.~Westbury, E.~Byrne, Z.~Chavis, A.~Furnari, R.~Girdhar,
  J.~Hamburger, H.~Jiang, M.~Liu, X.~Liu, et~al.
\newblock Ego4d: Around the world in 3,000 hours of egocentric video.
\newblock In \emph{IEEE Conf. Comput. Vis. Pattern Recog.}, pages 18995--19012,
  2022.

\bibitem[Greff et~al.(2022)Greff, Belletti, Beyer, Doersch, Du, Duckworth,
  Fleet, Gnanapragasam, Golemo, Herrmann, et~al.]{greff2022kubric}
K.~Greff, F.~Belletti, L.~Beyer, C.~Doersch, Y.~Du, D.~Duckworth, D.~J. Fleet,
  D.~Gnanapragasam, F.~Golemo, C.~Herrmann, et~al.
\newblock Kubric: A scalable dataset generator.
\newblock In \emph{IEEE Conf. Comput. Vis. Pattern Recog.}, pages 3749--3761,
  2022.

\bibitem[Gretton et~al.(2012)Gretton, Borgwardt, Rasch, Sch{\"o}lkopf, and
  Smola]{gretton2012kernel}
A.~Gretton, K.~M. Borgwardt, M.~J. Rasch, B.~Sch{\"o}lkopf, and A.~Smola.
\newblock A kernel two-sample test.
\newblock \emph{The Journal of Machine Learning Research}, 13\penalty0
  (1):\penalty0 723--773, 2012.

\bibitem[Gu et~al.(2024)Gu, Zhou, Wu, Yu, Liu, Zhao, Wu, Zhang, Shou, and
  Tang]{gu2024videoswap}
Y.~Gu, Y.~Zhou, B.~Wu, L.~Yu, J.-W. Liu, R.~Zhao, J.~Z. Wu, D.~J. Zhang, M.~Z.
  Shou, and K.~Tang.
\newblock Videoswap: Customized video subject swapping with interactive
  semantic point correspondence.
\newblock In \emph{IEEE Conf. Comput. Vis. Pattern Recog.}, pages 7621--7630,
  2024.

\bibitem[Gupta et~al.(2024)Gupta, Yu, Sohn, Gu, Hahn, Li, Essa, Jiang, and
  Lezama]{gupta2025photorealistic}
A.~Gupta, L.~Yu, K.~Sohn, X.~Gu, M.~Hahn, F.-F. Li, I.~Essa, L.~Jiang, and
  J.~Lezama.
\newblock Photorealistic video generation with diffusion models.
\newblock In \emph{Eur. Conf. Comput. Vis.}, pages 393--411. Springer, 2024.

\bibitem[He et~al.(2022)He, Yang, Zhang, Shan, and Chen]{he2022latent}
Y.~He, T.~Yang, Y.~Zhang, Y.~Shan, and Q.~Chen.
\newblock Latent video diffusion models for high-fidelity long video
  generation.
\newblock \emph{arXiv preprint arXiv:2211.13221}, 2022.

\bibitem[Hendrycks and Dietterich(2018)]{hendrycks2018benchmarking}
D.~Hendrycks and T.~Dietterich.
\newblock Benchmarking neural network robustness to common corruptions and
  perturbations.
\newblock In \emph{Int. Conf. Learn. Represent.}, 2018.

\bibitem[Hessel et~al.(2021)Hessel, Holtzman, Forbes, Bras, and
  Choi]{hessel2021clipscore}
J.~Hessel, A.~Holtzman, M.~Forbes, R.~L. Bras, and Y.~Choi.
\newblock Clipscore: A reference-free evaluation metric for image captioning.
\newblock \emph{Conf. on Emp. Meth. in Nat. Lang. Proc.}, 2021.

\bibitem[Ho et~al.(2022{\natexlab{a}})Ho, Chan, Saharia, Whang, Gao, Gritsenko,
  Kingma, Poole, Norouzi, Fleet, et~al.]{ho2022imagen}
J.~Ho, W.~Chan, C.~Saharia, J.~Whang, R.~Gao, A.~Gritsenko, D.~P. Kingma,
  B.~Poole, M.~Norouzi, D.~J. Fleet, et~al.
\newblock Imagen video: High definition video generation with diffusion models.
\newblock \emph{arXiv preprint arXiv:2210.02303}, 2022{\natexlab{a}}.

\bibitem[Ho et~al.(2022{\natexlab{b}})Ho, Salimans, Gritsenko, Chan, Norouzi,
  and Fleet]{ho2022video}
J.~Ho, T.~Salimans, A.~Gritsenko, W.~Chan, M.~Norouzi, and D.~J. Fleet.
\newblock Video diffusion models.
\newblock \emph{Adv. Neural Inform. Process. Syst.}, 35:\penalty0 8633--8646,
  2022{\natexlab{b}}.

\bibitem[Hong et~al.(2022)Hong, Ding, Zheng, Liu, and Tang]{hong2022cogvideo}
W.~Hong, M.~Ding, W.~Zheng, X.~Liu, and J.~Tang.
\newblock Cogvideo: Large-scale pretraining for text-to-video generation via
  transformers.
\newblock \emph{arXiv preprint arXiv:2205.15868}, 2022.

\bibitem[Hotshot-XL()]{hotshot}
Hotshot-XL.
\newblock Hotshot-xl.
\newblock \url{https://huggingface.co/hotshotco/Hotshot-XL.}

\bibitem[Huang et~al.(2024)Huang, He, Yu, Zhang, Si, Jiang, Zhang, Wu, Jin,
  Chanpaisit, et~al.]{huang2024vbench}
Z.~Huang, Y.~He, J.~Yu, F.~Zhang, C.~Si, Y.~Jiang, Y.~Zhang, T.~Wu, Q.~Jin,
  N.~Chanpaisit, et~al.
\newblock Vbench: Comprehensive benchmark suite for video generative models.
\newblock In \emph{IEEE Conf. Comput. Vis. Pattern Recog.}, pages 21807--21818,
  2024.

\bibitem[Jaegle et~al.(2021)Jaegle, Gimeno, Brock, Vinyals, Zisserman, and
  Carreira]{jaegle2021perceiver}
A.~Jaegle, F.~Gimeno, A.~Brock, O.~Vinyals, A.~Zisserman, and J.~Carreira.
\newblock Perceiver: General perception with iterative attention.
\newblock In \emph{Int. Conf. Mach. Learn.}, pages 4651--4664. PMLR, 2021.

\bibitem[Jaegle et~al.(2022)Jaegle, Borgeaud, Alayrac, Doersch, Ionescu, Ding,
  Koppula, Zoran, Brock, Shelhamer, et~al.]{jaegle2022perceiver}
A.~Jaegle, S.~Borgeaud, J.-B. Alayrac, C.~Doersch, C.~Ionescu, D.~Ding,
  S.~Koppula, D.~Zoran, A.~Brock, E.~Shelhamer, et~al.
\newblock {Perceiver IO}: A general architecture for structured inputs \&
  outputs.
\newblock In \emph{Int. Conf. Learn. Represent.}, 2022.

\bibitem[Khachatryan et~al.(2023)Khachatryan, Movsisyan, Tadevosyan, Henschel,
  Wang, Navasardyan, and Shi]{khachatryan2023text2video}
L.~Khachatryan, A.~Movsisyan, V.~Tadevosyan, R.~Henschel, Z.~Wang,
  S.~Navasardyan, and H.~Shi.
\newblock Text2video-zero: Text-to-image diffusion models are zero-shot video
  generators.
\newblock In \emph{Int. Conf. Comput. Vis.}, pages 15954--15964, 2023.

\bibitem[Kim et~al.(2024)Kim, Kim, and Yoo]{kim2024stream}
P.~J. Kim, S.~Kim, and J.~Yoo.
\newblock Stream: Spatio-temporal evaluation and analysis metric for video
  generative models.
\newblock In \emph{Int. Conf. Learn. Represent.}, 2024.

\bibitem[Kingma and Ba(2015)]{adam}
D.~P. Kingma and J.~Ba.
\newblock Adam: A method for stochastic optimization.
\newblock In \emph{Int. Conf. Learn. Represent.}, 2015.

\bibitem[Krizhevsky et~al.(2012)Krizhevsky, Sutskever, and
  Hinton]{krizhevsky2012imagenet}
A.~Krizhevsky, I.~Sutskever, and G.~E. Hinton.
\newblock Imagenet classification with deep convolutional neural networks.
\newblock \emph{Adv. Neural Inform. Process. Syst.}, 25, 2012.

\bibitem[Lai et~al.(2018)Lai, Huang, Wang, Shechtman, Yumer, and
  Yang]{lai2018learning}
W.-S. Lai, J.-B. Huang, O.~Wang, E.~Shechtman, E.~Yumer, and M.-H. Yang.
\newblock Learning blind video temporal consistency.
\newblock In \emph{Eur. Conf. Comput. Vis.}, pages 170--185, 2018.

\bibitem[Lin et~al.(2020)Lin, Wu, Peri, Fu, Jiang, and Ahn]{lin2020improving}
Z.~Lin, Y.-F. Wu, S.~Peri, B.~Fu, J.~Jiang, and S.~Ahn.
\newblock Improving generative imagination in object-centric world models.
\newblock In \emph{Int. Conf. Mach. Learn.}, pages 6140--6149. PMLR, 2020.

\bibitem[Liu et~al.(2024{\natexlab{a}})Liu, Qu, Yan, Zeng, Wang, and
  Liao]{liu2024frechet}
J.~Liu, Y.~Qu, Q.~Yan, X.~Zeng, L.~Wang, and R.~Liao.
\newblock Fr\'echet video motion distance: A metric for evaluating motion
  consistency in videos.
\newblock \emph{arXiv preprint arXiv:2407.16124}, 2024{\natexlab{a}}.

\bibitem[Liu et~al.(2024{\natexlab{b}})Liu, Cao, Wu, Mao, Gu, Zhao, Keppo,
  Shan, and Shou]{liu2024dynvideo}
J.-W. Liu, Y.-P. Cao, J.~Z. Wu, W.~Mao, Y.~Gu, R.~Zhao, J.~Keppo, Y.~Shan, and
  M.~Z. Shou.
\newblock Dynvideo-e: Harnessing dynamic nerf for large-scale motion-and
  view-change human-centric video editing.
\newblock In \emph{IEEE Conf. Comput. Vis. Pattern Recog.}, pages 7664--7674,
  2024{\natexlab{b}}.

\bibitem[Liu et~al.(2024{\natexlab{c}})Liu, Cun, Liu, Wang, Zhang, Chen, Liu,
  Zeng, Chan, and Shan]{liu2024evalcrafter}
Y.~Liu, X.~Cun, X.~Liu, X.~Wang, Y.~Zhang, H.~Chen, Y.~Liu, T.~Zeng, R.~Chan,
  and Y.~Shan.
\newblock Evalcrafter: Benchmarking and evaluating large video generation
  models.
\newblock In \emph{IEEE Conf. Comput. Vis. Pattern Recog.}, pages 22139--22149,
  2024{\natexlab{c}}.

\bibitem[Liu et~al.(2024{\natexlab{d}})Liu, Li, Ren, Gao, Li, Chen, Sun, and
  Hou]{liu2024fetv}
Y.~Liu, L.~Li, S.~Ren, R.~Gao, S.~Li, S.~Chen, X.~Sun, and L.~Hou.
\newblock Fetv: A benchmark for fine-grained evaluation of open-domain
  text-to-video generation.
\newblock \emph{Adv. Neural Inform. Process. Syst.}, 36, 2024{\natexlab{d}}.

\bibitem[Luo et~al.(2025)Luo, Favero, Hao~Luo, Jolicoeur-Martineau, and
  Pal]{luo2024beyond}
G.~Y. Luo, G.~M. Favero, Z.~Hao~Luo, A.~Jolicoeur-Martineau, and C.~Pal.
\newblock Beyond fvd: Enhanced evaluation metrics for video generation quality.
\newblock In \emph{Int. Conf. Learn. Represent.}, 2025.

\bibitem[Luo et~al.(2023)Luo, Chen, Zhang, Huang, Wang, Shen, Zhao, Zhou, and
  Tan]{luo2023videofusion}
Z.~Luo, D.~Chen, Y.~Zhang, Y.~Huang, L.~Wang, Y.~Shen, D.~Zhao, J.~Zhou, and
  T.~Tan.
\newblock Videofusion: Decomposed diffusion models for high-quality video
  generation.
\newblock \emph{arXiv preprint arXiv:2303.08320}, 2023.

\bibitem[Mayer et~al.(2016)Mayer, Ilg, Hausser, Fischer, Cremers, Dosovitskiy,
  and Brox]{mayer2016large}
N.~Mayer, E.~Ilg, P.~Hausser, P.~Fischer, D.~Cremers, A.~Dosovitskiy, and
  T.~Brox.
\newblock A large dataset to train convolutional networks for disparity,
  optical flow, and scene flow estimation.
\newblock In \emph{IEEE Conf. Comput. Vis. Pattern Recog.}, pages 4040--4048,
  2016.

\bibitem[MoonValley()]{moonvalley}
MoonValley.
\newblock Moonvalley.
\newblock \url{https://www.moonvalley.com/}.

\bibitem[Otani et~al.(2023)Otani, Togashi, Sawai, Ishigami, Nakashima, Rahtu,
  Heikkil{"a}, and Satoh]{otani2023toward}
M.~Otani, R.~Togashi, Y.~Sawai, R.~Ishigami, Y.~Nakashima, E.~Rahtu,
  J.~Heikkil{"a}, and S.~Satoh.
\newblock Toward verifiable and reproducible human evaluation for text-to-image
  generation.
\newblock In \emph{IEEE Conf. Comput. Vis. Pattern Recog.}, pages 14277--14286,
  2023.

\bibitem[Pictures()]{floor33}
F.~Pictures.
\newblock Floor33 pictures.
\newblock \url{https://www.morphstudio.com/}.

\bibitem[Pika()]{pika}
Pika.
\newblock Pika — pika.art.
\newblock \url{https://pika.art/}.

\bibitem[Radford et~al.(2021)Radford, Kim, Hallacy, Ramesh, Goh, Agarwal,
  Sastry, Askell, Mishkin, Clark, et~al.]{radford2021learning}
A.~Radford, J.~W. Kim, C.~Hallacy, A.~Ramesh, G.~Goh, S.~Agarwal, G.~Sastry,
  A.~Askell, P.~Mishkin, J.~Clark, et~al.
\newblock Learning transferable visual models from natural language
  supervision.
\newblock In \emph{Int. Conf. Mach. Learn.}, pages 8748--8763. PMLR, 2021.

\bibitem[Sajjadi et~al.(2022)Sajjadi, Meyer, Pot, Bergmann, Greff, Radwan,
  Vora, Lu{\v{c}}i{\'c}, Duckworth, Dosovitskiy, et~al.]{sajjadi2022scene}
M.~S. Sajjadi, H.~Meyer, E.~Pot, U.~Bergmann, K.~Greff, N.~Radwan, S.~Vora,
  M.~Lu{\v{c}}i{\'c}, D.~Duckworth, A.~Dosovitskiy, et~al.
\newblock Scene representation transformer: Geometry-free novel view synthesis
  through set-latent scene representations.
\newblock In \emph{IEEE Conf. Comput. Vis. Pattern Recog.}, pages 6229--6238,
  2022.

\bibitem[Saxena et~al.(2024)Saxena, Herrmann, Hur, Kar, Norouzi, Sun, and
  Fleet]{saxena2024surprising}
S.~Saxena, C.~Herrmann, J.~Hur, A.~Kar, M.~Norouzi, D.~Sun, and D.~J. Fleet.
\newblock The surprising effectiveness of diffusion models for optical flow and
  monocular depth estimation.
\newblock \emph{Adv. Neural Inform. Process. Syst.}, 36, 2024.

\bibitem[Simonyan and Zisserman(2014)]{simonyan2014very}
K.~Simonyan and A.~Zisserman.
\newblock Very deep convolutional networks for large-scale image recognition.
\newblock \emph{arXiv preprint arXiv:1409.1556}, 2014.

\bibitem[Singer et~al.(2022)Singer, Polyak, Hayes, Yin, An, Zhang, Hu, Yang,
  Ashual, Gafni, et~al.]{singer2022make}
U.~Singer, A.~Polyak, T.~Hayes, X.~Yin, J.~An, S.~Zhang, Q.~Hu, H.~Yang,
  O.~Ashual, O.~Gafni, et~al.
\newblock Make-a-video: Text-to-video generation without text-video data.
\newblock \emph{arXiv preprint arXiv:2209.14792}, 2022.

\bibitem[Soomro et~al.(2012)Soomro, Zamir, and Shah]{soomro2012ucf101}
K.~Soomro, A.~R. Zamir, and M.~Shah.
\newblock {UCF101:} {A} dataset of 101 human actions classes from videos in the
  wild.
\newblock \emph{arXiv preprint arXiv:1212.0402}, 2012.

\bibitem[Sterling()]{zeroscope}
S.~Sterling.
\newblock Zeroscope.
\newblock \url{https://huggingface.co/cerspense/zeroscope_v2_576w}.

\bibitem[Sun et~al.(2021)Sun, Vlasic, Herrmann, Jampani, Krainin, Chang, Zabih,
  Freeman, and Liu]{sun2021autoflow}
D.~Sun, D.~Vlasic, C.~Herrmann, V.~Jampani, M.~Krainin, H.~Chang, R.~Zabih,
  W.~T. Freeman, and C.~Liu.
\newblock Autoflow: Learning a better training set for optical flow.
\newblock In \emph{IEEE Conf. Comput. Vis. Pattern Recog.}, pages 10093--10102,
  2021.

\bibitem[Sun et~al.(2022)Sun, Herrmann, Reda, Rubinstein, Fleet, and
  Freeman]{sun2022disentangling}
D.~Sun, C.~Herrmann, F.~Reda, M.~Rubinstein, D.~J. Fleet, and W.~T. Freeman.
\newblock Disentangling architecture and training for optical flow.
\newblock In \emph{Eur. Conf. Comput. Vis.}, pages 165--182. Springer, 2022.

\bibitem[Teed and Deng(2020)]{teed2020raft}
Z.~Teed and J.~Deng.
\newblock Raft: Recurrent all-pairs field transforms for optical flow.
\newblock In \emph{Eur. Conf. Comput. Vis.}, pages 402--419. Springer, 2020.

\bibitem[Tong et~al.(2022)Tong, Song, Wang, and Wang]{tong2022videomae}
Z.~Tong, Y.~Song, J.~Wang, and L.~Wang.
\newblock Videomae: Masked autoencoders are data-efficient learners for
  self-supervised video pre-training.
\newblock \emph{Adv. Neural Inform. Process. Syst.}, 35:\penalty0 10078--10093,
  2022.

\bibitem[Unterthiner et~al.(2018)Unterthiner, Van~Steenkiste, Kurach, Marinier,
  Michalski, and Gelly]{unterthiner2018towards}
T.~Unterthiner, S.~Van~Steenkiste, K.~Kurach, R.~Marinier, M.~Michalski, and
  S.~Gelly.
\newblock Towards accurate generative models of video: A new metric \&
  challenges.
\newblock \emph{arXiv preprint arXiv:1812.01717}, 2018.

\bibitem[van Steenkiste et~al.(2024)van Steenkiste, Zoran, Yang, Rubanova,
  Kabra, Doersch, Gokay, Heyward, Pot, Greff, Hudson, Keck, Carreira,
  Dosovitskiy, Sajjadi, and Kipf]{vanSteenkiste2024moving}
S.~van Steenkiste, D.~Zoran, Y.~Yang, Y.~Rubanova, R.~Kabra, C.~Doersch,
  D.~Gokay, J.~Heyward, E.~Pot, K.~Greff, D.~A. Hudson, T.~A. Keck,
  J.~Carreira, A.~Dosovitskiy, M.~S.~M. Sajjadi, and T.~Kipf.
\newblock Moving off-the-grid: Scene-grounded video representations.
\newblock In \emph{Adv. Neural Inform. Process. Syst.}, 2024.

\bibitem[Vaswani et~al.(2017)Vaswani, Shazeer, Parmar, Uszkoreit, Jones, Gomez,
  Kaiser, and Polosukhin]{transformer}
A.~Vaswani, N.~Shazeer, N.~Parmar, J.~Uszkoreit, L.~Jones, A.~N. Gomez,
  {\L}.~Kaiser, and I.~Polosukhin.
\newblock Attention is all you need.
\newblock \emph{Adv. Neural Inform. Process. Syst.}, 30, 2017.

\bibitem[Venkataramanan et~al.(2024)Venkataramanan, Rizve, Carreira, Asano, and
  Avrithis]{venkataramanan2024imagenet}
S.~Venkataramanan, M.~N. Rizve, J.~Carreira, Y.~Asano, and Y.~Avrithis.
\newblock Is imagenet worth 1 video? learning strong image encoders from 1 long
  unlabelled video.
\newblock In \emph{Int. Conf. Learn. Represent.}, pages 1--21, 2024.

\bibitem[Villegas et~al.(2022)Villegas, Babaeizadeh, Kindermans, Moraldo,
  Zhang, Saffar, Castro, Kunze, and Erhan]{villegas2022phenaki}
R.~Villegas, M.~Babaeizadeh, P.-J. Kindermans, H.~Moraldo, H.~Zhang, M.~T.
  Saffar, S.~Castro, J.~Kunze, and D.~Erhan.
\newblock Phenaki: Variable length video generation from open domain textual
  descriptions.
\newblock In \emph{Int. Conf. Learn. Represent.}, 2022.

\bibitem[Wang et~al.(2023{\natexlab{a}})Wang, Yuan, Chen, Zhang, Wang, and
  Zhang]{wang2023modelscope}
J.~Wang, H.~Yuan, D.~Chen, Y.~Zhang, X.~Wang, and S.~Zhang.
\newblock Modelscope text-to-video technical report.
\newblock \emph{arXiv preprint arXiv:2308.06571}, 2023{\natexlab{a}}.

\bibitem[Wang et~al.(2023{\natexlab{b}})Wang, Huang, Zhao, Tong, He, Wang,
  Wang, and Qiao]{wang2023videomae}
L.~Wang, B.~Huang, Z.~Zhao, Z.~Tong, Y.~He, Y.~Wang, Y.~Wang, and Y.~Qiao.
\newblock Videomae v2: Scaling video masked autoencoders with dual masking.
\newblock In \emph{IEEE Conf. Comput. Vis. Pattern Recog.}, pages 14549--14560,
  2023{\natexlab{b}}.

\bibitem[Wang et~al.(2020)Wang, Zhu, Wang, Hu, Qiu, Wang, Hu, Kapoor, and
  Scherer]{wang2020tartanair}
W.~Wang, D.~Zhu, X.~Wang, Y.~Hu, Y.~Qiu, C.~Wang, Y.~Hu, A.~Kapoor, and
  S.~Scherer.
\newblock Tartanair: A dataset to push the limits of visual slam.
\newblock In \emph{IEEE/RSJ Int. Conf. Intell. Robots Syst.}, pages 4909--4916.
  IEEE, 2020.

\bibitem[Wang et~al.(2017)Wang, Long, Wang, Gao, and Yu]{wang2017predrnn}
Y.~Wang, M.~Long, J.~Wang, Z.~Gao, and P.~S. Yu.
\newblock Predrnn: Recurrent neural networks for predictive learning using
  spatiotemporal lstms.
\newblock \emph{Adv. Neural Inform. Process. Syst.}, 30, 2017.

\bibitem[Wang et~al.(2023{\natexlab{c}})Wang, Chen, Ma, Zhou, Huang, Wang,
  Yang, He, Yu, Yang, et~al.]{wang2023lavie}
Y.~Wang, X.~Chen, X.~Ma, S.~Zhou, Z.~Huang, Y.~Wang, C.~Yang, Y.~He, J.~Yu,
  P.~Yang, et~al.
\newblock Lavie: High-quality video generation with cascaded latent diffusion
  models.
\newblock \emph{arXiv preprint arXiv:2309.15103}, 2023{\natexlab{c}}.

\bibitem[Wang et~al.(2004)Wang, Bovik, Sheikh, and Simoncelli]{wang2004image}
Z.~Wang, A.~C. Bovik, H.~R. Sheikh, and E.~P. Simoncelli.
\newblock Image quality assessment: from error visibility to structural
  similarity.
\newblock \emph{IEEE Trans. Image Process.}, 13\penalty0 (4):\penalty0
  600--612, 2004.

\bibitem[Whitney et~al.(2023)Whitney, Lopez-Guevara, Pfaff, Rubanova, Kipf,
  Stachenfeld, and Allen]{whitney2023learning}
W.~F. Whitney, T.~Lopez-Guevara, T.~Pfaff, Y.~Rubanova, T.~Kipf,
  K.~Stachenfeld, and K.~R. Allen.
\newblock Learning 3d particle-based simulators from rgb-d videos.
\newblock \emph{Int. Conf. Learn. Represent.}, 2023.

\bibitem[Wu et~al.(2021)Wu, Huang, Zhang, Li, Ji, Yang, Sapiro, and
  Duan]{wu2021godiva}
C.~Wu, L.~Huang, Q.~Zhang, B.~Li, L.~Ji, F.~Yang, G.~Sapiro, and N.~Duan.
\newblock Godiva: Generating open-domain videos from natural descriptions.
\newblock \emph{arXiv preprint arXiv:2104.14806}, 2021.

\bibitem[Wu et~al.(2022{\natexlab{a}})Wu, Liang, Ji, Yang, Fang, Jiang, and
  Duan]{wu2022nuwa}
C.~Wu, J.~Liang, L.~Ji, F.~Yang, Y.~Fang, D.~Jiang, and N.~Duan.
\newblock N{"u}wa: Visual synthesis pre-training for neural visual world
  creation.
\newblock In \emph{Eur. Conf. Comput. Vis.}, pages 720--736. Springer,
  2022{\natexlab{a}}.

\bibitem[Wu et~al.(2022{\natexlab{b}})Wu, Liao, Chen, Hou, Wang, Sun, Yan, and
  Lin]{wu2022disentangling}
H.~Wu, L.~Liao, C.~Chen, J.~Hou, A.~Wang, W.~Sun, Q.~Yan, and W.~Lin.
\newblock Disentangling aesthetic and technical effects for video quality
  assessment of user generated content.
\newblock \emph{arXiv preprint arXiv:2211.04894}, 2\penalty0 (5):\penalty0 6,
  2022{\natexlab{b}}.

\bibitem[Wu et~al.(2023)Wu, Ge, Wang, Lei, Gu, Shi, Hsu, Shan, Qie, and
  Shou]{wu2023tune}
J.~Z. Wu, Y.~Ge, X.~Wang, S.~W. Lei, Y.~Gu, Y.~Shi, W.~Hsu, Y.~Shan, X.~Qie,
  and M.~Z. Shou.
\newblock Tune-a-video: One-shot tuning of image diffusion models for
  text-to-video generation.
\newblock In \emph{Int. Conf. Comput. Vis.}, 2023.

\bibitem[Wu et~al.(2024)Wu, Fang, Wu, Wang, Ge, Cun, Zhang, Liu, Gu, Zhao,
  et~al.]{wu2024towards}
J.~Z. Wu, G.~Fang, H.~Wu, X.~Wang, Y.~Ge, X.~Cun, D.~J. Zhang, J.-W. Liu,
  Y.~Gu, R.~Zhao, et~al.
\newblock Towards a better metric for text-to-video generation.
\newblock \emph{arXiv preprint arXiv:2401.07781}, 2024.

\bibitem[Xiong et~al.(2020)Xiong, Yang, He, Zheng, Zheng, Xing, Zhang, Lan,
  Wang, and Liu]{xiong2020layer}
R.~Xiong, Y.~Yang, D.~He, K.~Zheng, S.~Zheng, C.~Xing, H.~Zhang, Y.~Lan,
  L.~Wang, and T.~Liu.
\newblock On layer normalization in the transformer architecture.
\newblock In \emph{Int. Conf. Mach. Learn.}, pages 10524--10533. PMLR, 2020.

\bibitem[Yang et~al.(2024)Yang, Teng, Zheng, Ding, Huang, Xu, Yang, Hong,
  Zhang, Feng, et~al.]{yang2024cogvideox}
Z.~Yang, J.~Teng, W.~Zheng, M.~Ding, S.~Huang, J.~Xu, Y.~Yang, W.~Hong,
  X.~Zhang, G.~Feng, et~al.
\newblock Cogvideox: Text-to-video diffusion models with an expert transformer.
\newblock \emph{arXiv preprint arXiv:2408.06072}, 2024.

\bibitem[Zhang et~al.(2024)Zhang, Wu, Liu, Zhao, Ran, Gu, Gao, and
  Shou]{zhang2024show}
D.~J. Zhang, J.~Z. Wu, J.-W. Liu, R.~Zhao, L.~Ran, Y.~Gu, D.~Gao, and M.~Z.
  Shou.
\newblock Show-1: Marrying pixel and latent diffusion models for text-to-video
  generation.
\newblock \emph{International Journal of Computer Vision}, pages 1--15, 2024.

\bibitem[Zhang et~al.(2018)Zhang, Isola, Efros, Shechtman, and
  Wang]{zhang2018perceptual}
R.~Zhang, P.~Isola, A.~A. Efros, E.~Shechtman, and O.~Wang.
\newblock The unreasonable effectiveness of deep features as a perceptual
  metric.
\newblock In \emph{IEEE Conf. Comput. Vis. Pattern Recog.}, 2018.

\bibitem[Zheng et~al.(2023)Zheng, Harley, Shen, Wetzstein, and
  Guibas]{zheng2023pointodyssey}
Y.~Zheng, A.~W. Harley, B.~Shen, G.~Wetzstein, and L.~J. Guibas.
\newblock Pointodyssey: A large-scale synthetic dataset for long-term point
  tracking.
\newblock In \emph{Int. Conf. Comput. Vis.}, pages 19855--19865, 2023.

\bibitem[Zheng et~al.()Zheng, Peng, Yang, Shen, Li, Liu, Zhou, Li, and
  You]{zheng1open}
Z.~Zheng, X.~Peng, T.~Yang, C.~Shen, S.~Li, H.~Liu, Y.~Zhou, T.~Li, and Y.~You.
\newblock Open-sora: Democratizing efficient video production for all, march
  2024.
\newblock \emph{URL https://github. com/hpcaitech/Open-Sora}, 1\penalty0
  (3):\penalty0 4.

\bibitem[Zhou et~al.(2022)Zhou, Wang, Yan, Lv, Zhu, and
  Feng]{zhou2022magicvideo}
D.~Zhou, W.~Wang, H.~Yan, W.~Lv, Y.~Zhu, and J.~Feng.
\newblock Magicvideo: Efficient video generation with latent diffusion models.
\newblock \emph{arXiv preprint arXiv:2211.11018}, 2022.

\end{thebibliography}
